\documentclass{article} 
\usepackage{iclr2027_conference,times}

\usepackage{amsmath,amsfonts,bm}

\newcommand{\SPN}{\mathrm{span}}
\newcommand{\ip}[2]{\left\langle #1,#2\right\rangle}
\newcommand{\norm}[1]{\left\lVert #1\right\rVert}
\newcommand{\msign}{\mathrm{msign}}

\newcommand{\nuc}[1]{\left\lVert #1\right\rVert_*}

\def\eqref#1{\textcolor{blue}{Eq.}~(\ref{#1})}

\def\1{\bm{1}}

\def\cc{{c_\beta}}

\def\rvm{{\mathbf{m}}}

\def\rvv{{\mathbf{v}}}

\newcommand{\Ex}[1]{\mathbb{E}\left[{#1}\right]}

\DeclareMathAlphabet{\mathsfit}{\encodingdefault}{\sfdefault}{m}{sl}
\SetMathAlphabet{\mathsfit}{bold}{\encodingdefault}{\sfdefault}{bx}{n}

\newcommand{\E}{\mathbb{E}}

\newcommand{\R}{\mathbb{R}}

\DeclareMathOperator*{\argmax}{arg\,max}

\newcommand{\tr}{\mathrm{tr}}

\newtheorem{thm}{Theorem}

\newtheorem{lem}{Lemma}
\newtheorem{rem}{Remark}

\usepackage{graphicx}
\newtheorem{assumption}{Assumption}
\usepackage{amssymb}
\usepackage{algorithm,algorithmic}
\usepackage{hyperref}
\usepackage{url}
\usepackage[capitalize,noabbrev]{cleveref}
\hypersetup{
    colorlinks,
    breaklinks,
    urlcolor = black,
    linkcolor = blue,
    citecolor = blue,
}
\usepackage{wrapfig}
\usepackage{multirow}
\usepackage{subfigure}
\title{Memory-Efficient Subspace Optimization with Residual Correction: From Momentum SGD to Muon}

\author{
  Sifan Yang\textsuperscript{\rm 1,2,3},~Hongyi Tao\textsuperscript{\rm 1,2},~Dingzhi Yu\textsuperscript{\rm 1,2},~Haomin Bai\textsuperscript{\rm 1,2},~Wei Jiang\textsuperscript{\rm 4,1},~Silin Gao\textsuperscript{\rm 3},\\
  ~\textbf{Pi Bu}\textsuperscript{\rm 3},~\textbf{Lijun Zhang}\textsuperscript{\rm 1,2,}\thanks{Lijun Zhang is the corresponding author.}\\
  \textsuperscript{\rm 1}State Key Laboratory of Novel Software Technology, Nanjing University \\
  \textsuperscript{\rm 2}School of Artificial Intelligence, Nanjing University \\
   \textsuperscript{\rm 3}Alibaba Token Hub, Alibaba Group\\
   \textsuperscript{\rm 4}School of Computer Science and Engineering, Nanjing University of Science and Technology
}
\iclrfinalcopy
\begin{document}

\maketitle
\lhead{Preprint}

\begin{abstract}%
Memory-efficient subspace optimization has emerged as a practical approach to reducing the memory cost of LLM training while preserving optimization performance. Previous algorithms typically rely on the leading singular subspace of stochastic gradients, yet the gradient noise may dominate the subspace in later training stages and can prevent convergence. To overcome this limitation, we adopt a residual correction framework, which maintains the optimizer states in the subspace and adds the orthogonal residual gradient to the parameter update. 
We first integrate this scheme into momentum SGD to develop SMRC, which attains an $O(T^{-1/4})$ convergence rate for a fixed orthonormal basis, matching that of full-space momentum SGD.
Our analysis introduces a \emph{subspace virtual iterate} to resolve the coupling between the momentum and the residual correction. Furthermore, we extend SMRC to a multi-stage variant which periodically recomputes bases to track the evolving dominant subspace and retains the same rate.
Since each basis refresh breaks the consistency of the virtual iterates, we devise a \emph{boundary correction} approach that aligns them across stages. 
Beyond momentum SGD, we incorporate our framework into Muon and propose PMRC, which applies the matrix sign to the projected momentum. Through a \emph{weighted momentum tracking} analysis that accounts for the residual gradient, we establish an $O(T^{-1/4})$ convergence rate for PMRC, which matches the rate of full-space Muon.  We also design a multi-stage PMRC with periodically refreshed bases, which maintains a convergence rate of the same order. Extensive experimental results confirm the effectiveness and efficiency of our algorithms.
\end{abstract}
\section{Introduction}
Large language models (LLMs) \citep{Transformer,GPT3,GPT4,Qwen,DeepSeekR1,KimiK2} have demonstrated strong capabilities in dialogue \citep{LaMDA}, language understanding \citep{PaLM}, mathematical reasoning \citep{Minerva} and code generation \citep{Codex}.  As model scale grows, training imposes substantial computational and memory demands \citep{GPT3,PaLM,LLaMA2}. The widely used AdamW optimizer \citep{Adam,AdamW} further increases memory consumption, since its first- and second-moment estimates are each as large as the model itself. To alleviate the memory overhead, recent work has explored different forms of memory-efficient training, ranging from restricting the set of trainable parameters \citep{Houlsby2019,BitFit,LoRA,AdaLoRA,LISA} to maintaining optimizer states in low-dimensional subspaces \citep{GaLore,GoLore,Fira,pan2025gum}.  
In this work,  we focus on memory-efficient subspace optimization, which reduces optimization memory by maintaining the optimizer states in a low-dimensional subspace. GaLore \citep{GaLore} is a representative method in this line of work. For a weight matrix $W_t\in\R^{m\times n}$, GaLore utilizes an orthonormal basis $P\in\R^{m\times r}$ with $r\ll\min\{m,n\}$ and projects the gradient $G_t\in\R^{m\times n}$ into the space $\R^{r \times n}$, where the optimizer maintains its states, reducing the memory complexity from $O(mn)$ to $O((m+n)r)$. To construct the projection matrix, GaLore computes the leading singular vectors of the  stochastic gradient to capture its dominant directions and periodically recomputes $P$ via SVD during training.

As shown by \citet{GoLore}, GaLore may fail to converge to a stationary point under standard stochastic assumptions. Since the projection basis is estimated from stochastic gradients, whose leading singular directions can be dominated by anisotropic noise \emph{rather than the true gradient} in later training stages, the selected subspace fails to capture gradient directions. While GoLore \citep{GoLore} replaces the SVD basis with random projections and establishes convergence under standard assumptions, its random subspaces do not exploit dominant gradient directions and are less effective in practice. Recent methods that retain gradient-dependent subspaces either require stronger assumptions  \citep{LDAdam} or offer no memory advantage in the worst case \citep{pan2025gum}. Therefore, there still \emph{remains a lack of convergence guarantees for memory-efficient subspace optimization with gradient-dependent dominant subspaces under standard assumptions.}

\textbf{Subspace Momentum SGD with Residual Correction (SMRC).} Inspired by \citet{Fira}, we adopt a residual correction framework, which maintains optimizer states in a low-dimensional subspace and adds the residual gradient in its \emph{orthogonal complement} to the update. We incorporate this framework into momentum SGD \citep{Momentum:SGD} and propose SMRC, which keeps momentum only in the projected space and reduces optimizer-state memory from $O(mn)$ to $O((m+n)r)$.
Although the projected momentum update and the residual correction lie in orthogonal subspaces, they remain coupled across iterations.  To overcome this difficulty, we devise a \emph{subspace virtual iterate} strategy that absorbs the momentum into an auxiliary trajectory. SMRC achieves an $O(T^{-1/4})$ convergence rate for any fixed orthonormal basis, matching that of full-space momentum SGD. Furthermore, since the dominant gradient subspace evolves during training,  practical methods periodically refresh the projection matrix $P$, and we accordingly extend SMRC to a multi-stage variant. A direct application of the single-stage analysis results in a mismatch between virtual iterates at stage boundaries. We design a \emph{boundary correction} approach that preserves the consistency of the virtual sequence. Our multi-stage SMRC attains the same rate as that of single-stage SMRC.

\textbf{Projected Muon with Residual Correction (PMRC).} Beyond momentum SGD, we apply the residual correction framework to Muon \citep{jordan2024muon}, an effective optimizer for LLM training \citep{MuonScalable,KimiK2}, and develop PMRC. It maintains momentum in the projected coordinates and applies the matrix sign to the low-dimensional momentum. This nonlinear dependence of the update on the momentum invalidates the subspace virtual-iterate construction used for SMRC.  To address this challenge, we present a \emph{weighted momentum tracking} analysis that jointly characterizes momentum accumulation and effects induced by the residual correction. The  matrix-sign step corresponds to steepest descent under the spectral norm and is characterized by the nuclear norm \citep{bernstein2024oldoptimizernewnorm}, whereas the residual gradient correction is measured in the Frobenius norm. We establish convergence under a mixed stationarity measure  and PMRC attains the same $O(T^{-1/4})$ rate as that of full-space Muon \citep{li2025muon}. Similarly, we generalize PMRC to a multi-stage setting with periodically refreshed projection bases that better capture the evolving dominant gradient subspace and improve practical performance. The multi-stage variant retains the same $O(T^{-1/4})$ convergence rate.

Finally, we evaluate our methods through extensive experiments. On synthetic problems, we examine the residual correction and the multi-stage basis refresh in SMRC, and the results show that the residual correction restores convergence while periodic refreshes track the evolving dominant gradient subspace. For practical LLM training, we develop a practical implementation of PMRC and evaluate it on LLaMA pretraining across \textbf{four} model scales from $60$M to $1.3$B parameters and on \textbf{six} supervised finetuning settings. Our PMRC consistently performs favorably against prior memory-efficient methods across different model scales and training settings.

\section{Related Work}
In this section, we briefly review related work on memory-efficient optimization and on Muon and its variants. Due to space limitations, we provide  additional discussion in Appendix~\ref{appendix:related}.

\subsection{Memory-efficient Optimization for LLM Training}
As model scale grows, full-parameter training becomes increasingly memory-intensive, since gradients and optimizer states impose substantial overhead. Parameter-efficient finetuning methods reduce this cost by restricting the update to a small set of  parameters.  LoRA \citep{LoRA} freezes the model weights and learns low-rank updates, which is effective for finetuning but limits the expressiveness of the update in pretraining \citep{DeltaTuning,ReLoRA}.   GaLore \citep{GaLore} instead reduces memory by maintaining optimizer states in a low-dimensional gradient subspace. Given a stochastic gradient $G_t\in\R^{m\times n}$, where we assume $m\leq n$ without loss of generality, GaLore uses an orthonormal basis $P_t\in\R^{m\times r}$ with $r\ll m$ to form the projected gradient $P_t^\top G_t\in\R^{r\times n}$. The optimizer states are maintained in the reduced coordinates and the resulting update is mapped back to the original parameter space through $P_t$. GaLore constructs $P_t$ from the top-$r$ left singular vectors of the  stochastic gradient. Since computing the top-$r$ SVD at every iteration is expensive, GaLore refreshes $P_t$ once every $\tau$ iterations while keeping it fixed between consecutive refreshes. As shown by \citet{GoLore}, the leading singular subspaces computed from stochastic gradients can be dominated by anisotropic noise rather than the true gradient, leaving informative gradient directions outside $\SPN(P_t)$ and preventing convergence under standard stochastic assumptions.

To address this limitation, GoLore \citep{GoLore} replaces the SVD basis with random projections. While GoLore establishes convergence guarantees under standard stochastic assumptions, its random subspaces do not preferentially capture dominant gradient directions and lead to slow convergence.  Fira \citep{Fira} follows a complementary approach by retaining the SVD-based subspace and augmenting the projected update with a norm-scaled correction of the discarded residual gradient. Although Fira performs well empirically, its residual scaling is primarily motivated by empirical observations and lacks a convergence guarantee. \citet{FRUGAL} and \citet{SubspaceMomentum} also maintain optimizer states in a subspace and apply an SGD-type update in the orthogonal complement. The analysis in \citet{FRUGAL} is developed for momentum SGD on selected coordinates combined with SGD on the remaining ones. Its convergence guarantee therefore does not directly cover the periodically refreshed SVD subspaces used in GaLore. The convergence rate of \citet{SubspaceMomentum} is established for a fixed basis under stronger assumptions and applies only to an unweighted residual update.


\subsection{Muon and its Variants}
Muon \citep{jordan2024muon} has demonstrated strong empirical performance in LLM training, outperforming AdamW while exhibiting stable scaling behavior \citep{KimiK2,MuonScalable}. For matrix-shaped parameters, Muon updates along the matrix sign of the momentum, the steepest-descent direction under the spectral norm, and approximates it efficiently through Newton--Schulz iterations. Despite these advantages, Muon still requires a momentum of the same size as the parameter matrix and  incurs the same $O(mn)$ optimizer-state memory cost as momentum SGD.

In recent years, a growing body of work has developed Muon variants to improve its memory efficiency \citep{QuantizedMuon,refael2025sumo,pan2025gum,LiMuon,sMuon}, and we provide a detailed discussion of these methods in Appendix~\ref{app:muon-related}. Among them, GUM \citep{pan2025gum} combines GaLore-style low-rank projection with layerwise sampling. At each sampling period, a subset of layers receives compensated full-rank updates, while the remaining layers receive projected low-rank ones. The two branches are reweighted according to their sampling probabilities. Since the full-rank branches are drawn independently across layers, all layers may be selected at the same time, in which case GUM offers no memory reduction over Muon.

\section{Notation and Assumptions}
In this section, we  present necessary preliminaries for our work.

\textbf{Notation.} We denote the optimization variable by $W\in\R^{m\times n}$ and assume $m\le n$ without loss of generality.  We use $I_r$ to denote the $r\times r$ identity matrix. For matrices $A,B\in\R^{m\times n}$, we define the trace inner product as $\ip{A}{B}=\tr(A^\top B)$ and denote the Frobenius, spectral and nuclear norms by $\norm{A}_F$, $\norm{A}_2$ and $\nuc{A}$, respectively. We refer to a matrix $P\in\R^{m\times r}$ with $P^\top P=I_r$ as an orthonormal basis. The matrix $PP^\top$ is the orthogonal projection matrix onto $\SPN(P)$ and $Q=I-PP^\top$ is the projection matrix onto $\SPN(P)^\perp$. All expectations are taken with respect to the randomness of the stochastic gradient oracle.

In the following, we introduce some standard assumptions in stochastic nonconvex optimization \citep{GoLore}.

\begin{assumption}[Smoothness]\label{ass:smooth}
\textnormal{The loss function $F(\cdot)$ is $L$-smooth, i.e.,
$\|\nabla F(W)-\nabla F(W')\|_F \le L\|W-W'\|_F,
\quad \forall W, W' \in \R^{m\times n}.$}
\end{assumption}

\begin{assumption}[Lower boundedness]\label{ass:lower:bound}
\textnormal{The loss function $F(\cdot)$ is lower bounded, i.e., $F_* = \inf_{W\in\R^{m\times n}} F(W) > -\infty$. We define $\Delta = F(W_1) - F_*$ as the initial optimality gap.}
\end{assumption}

\begin{assumption}[Stochastic gradients]\label{ass:var}
\textnormal{For any $W_t$, the stochastic gradient oracle returns $g(W_t)$ such that $\E\left[g(W_t)\right]=\nabla F(W_t)$ and $\E\left[\|g(W_t)-\nabla F(W_t)\|_F^2\right]\le \sigma^2.$}
\end{assumption}

\begin{assumption}[Orthonormal basis]\label{ass:predictable}
\textnormal{The basis $P_k\in\R^{m\times r}$ is determined by the optimization history prior to stage $k$ and satisfies $P_k^\top P_k=I_r$.}\end{assumption}

\section{Subspace Momentum SGD with Residual Correction (SMRC)}

In this section, we incorporate the residual correction framework into momentum SGD and develop SMRC. We first present the single-stage SMRC with a fixed orthonormal basis and establish its convergence guarantee. Building on this result, we then propose a multi-stage variant with periodically refreshed bases and show that it retains the same convergence rate.

\subsection{Single-Stage SMRC}
As discussed above, GaLore \citep{GaLore} discards the gradient component outside the subspace, which can prevent convergence under stochastic noise. GoLore \citep{GoLore} restores the convergence guarantee through random projections, yet its gradient-independent projection strategy leads to worse practical performance than GaLore.  To resolve this dilemma, we adopt a residual correction framework inspired by \cite{Fira}, which maintains the optimizer states in the low-rank subspace and adds the orthogonal residual gradient to the parameter update.

\begin{algorithm}[t]
\caption{Single-Stage SMRC}\label{alg:single}
\textbf{Input}: initial iterate $W_1$, $P \in \R^{m\times r}$, $\eta$, $\beta\in(0,1)$, $\gamma\in(0,1]$
\begin{algorithmic}[1]
\STATE Initialize momentum: $\rvm_{0}=\textbf{0}$
\FOR{$t=1,\dots,T$}
  \STATE Query the stochastic gradient $G_t=g(W_t)$
  \STATE Compute $\rvm_t = (1-\beta)\rvm_{t-1}+\beta P^\top G_t$
  \STATE Update $W_{t+1}= W_t-\eta\big(P\rvm_t+\gamma (I-PP^\top) G_t\big)$
\ENDFOR
\end{algorithmic}
\end{algorithm}

In this section, we incorporate the residual correction framework into momentum SGD, a widely used method, and develop SMRC, as summarized in Algorithm~\ref{alg:single}. At each iteration $t$, SMRC queries a stochastic gradient $G_t$, projects it onto the subspace and maintains the low-dimensional momentum $\rvm_t=(1-\beta)\rvm_{t-1}+\beta P^\top G_t\in\R^{r\times n}$, where $P\in\R^{m\times r}$ is a fixed orthonormal basis and $\beta\in(0,1)$ is the momentum parameter. When $P$ consists of the top-$r$ left singular vectors of the stochastic gradient, the projection $P^\top G_t$ retains the components along the estimated dominant directions, and the momentum is mapped back to the original parameter space through $P\rvm_t$. However, as pointed out by \citet{GoLore}, this subspace may primarily reflect the noise and miss informative gradient directions once the stochastic noise dominates the true gradient. To recover these directions, we incorporate the residual gradient into the update, i.e., $W_{t+1}=W_t-\eta(P\rvm_t+\gamma(I-PP^\top)G_t)$, where the parameter $\gamma\in(0,1]$ controls the weight of the residual gradient.

We next outline the key technical difficulty introduced by the residual correction, deferring the detailed argument to Appendix~\ref{sec:Orthogonal}. Since the two update directions are orthogonal, a natural approach is to analyze the projected and residual terms separately. However, this direct decomposition is insufficient, as the residual correction directly affects the iterate, which alters subsequent gradients that are projected and accumulated in the momentum, leaving the two components coupled across iterations. To overcome this challenge, we develop a subspace virtual iterate approach, which introduces the auxiliary sequence $U_t = W_t-\frac{(1-\beta)\eta}{\beta}P\rvm_{t-1}$ and $U_1 = W_1$. This sequence is used only in the analysis and is not computed by SMRC. The virtual iterate incorporates the projected momentum into an auxiliary trajectory, thereby decoupling it from the residual correction term.  Formally, we establish the theoretical guarantee for SMRC.

\begin{thm}[Single-stage SMRC] \label{thm:1} Define $Q=I-PP^\top$ and $\cc=\frac{1-\beta}{\beta}$. Under Assumptions~\ref{ass:smooth}, \ref{ass:lower:bound} and~\ref{ass:var}, for a matrix $P\in\R^{m\times r}$ with $P^\top P=I_r$ and by setting $\eta = \min \left\{\frac{\gamma}{2L},\frac{\gamma}{2L\cc}, \frac{1}{L\sqrt{T}}\right\}, \beta \in (0,1), \gamma \in (0,1]$, Algorithm \ref{alg:single} ensures
\begin{equation*}
    \frac{1}{T}\sum_{t=1}^{T}\Ex{\big\|\nabla F(W_t)\big\|_F}  = O(T^{-1/4}).
\end{equation*}
\end{thm}

\begin{rem}\textnormal{
For any $\beta\in(0,1),\gamma\in(0,1]$ and the step size prescribed in Theorem~\ref{thm:1}, SMRC attains an $O(T^{-1/4})$ convergence rate, which is standard for stochastic nonconvex optimization and matches that of full-space momentum SGD, while storing only a low-rank momentum.}
\end{rem}

\begin{rem}
\textnormal{
SMRC maintains the momentum $\rvm_t\in\R^{r\times n}$ in the projected space rather than in the ambient space $\R^{m\times n}$. The residual correction $(I-PP^\top)G_t$ can be computed directly from the stochastic gradient and needs no additional optimizer state \citep{Fira}.  Storing the basis $P\in\R^{m\times r}$ and the momentum $\rvm_t$ only requires $O((m+n)r)$ memory, which is smaller than the $O(mn)$ memory of full-space momentum SGD.  
}
\end{rem}

\begin{algorithm}[t]
\caption{Multi-Stage SMRC}\label{alg:multi}
\textbf{Input}: initial iterate $W_{1,1}$, $P_1$, $\eta$, $\beta\in(0,1)$, $\gamma\in(0,1]$, $K$, $\{T_k\}_{k=1}^K$
\begin{algorithmic}[1]
\FOR{$k=1,\dots,K$}
  \STATE Set momentum $\rvm_{k,0}=\textbf{0}$.
  \FOR{$t=1,\dots,T_k$}
    \STATE Query the stochastic gradient $G_{k,t}=g(W_{k,t})$
    \STATE Calculate $\rvm_{k,t}=(1-\beta)\rvm_{k,t-1}+\beta P_k^\top G_{k,t}$
    \STATE Update $W_{k,t+1}=W_{k,t}-\eta\big(P_k \rvm_{k,t}+\gamma(I-P_kP_k^\top)G_{k,t}\big)$
  \ENDFOR
  \IF{$k<K$}
    \STATE Set $W_{k+1,1}=W_{k,T_k+1}-\frac{(1-\beta)\eta}{\beta}P_k \rvm_{k,T_k}$  and update $P_{k+1}$ 
  \ENDIF
\ENDFOR
\end{algorithmic}
\end{algorithm}
\subsection{Multi-stage SMRC}\label{sec:smrc-multi}
Although the single-stage analysis accommodates any fixed orthonormal basis $P$, practical subspace methods \citep{GaLore,pan2025gum} typically construct $P$ from the dominant singular subspace of the stochastic gradient so that the projected update retains informative gradient directions. Since the dominant singular subspace of the gradient varies throughout optimization, practical subspace methods periodically update the projection basis to capture the evolving geometry of the loss landscape. Following the periodic refresh strategy of GaLore \citep{GaLore}, we propose the multi-stage SMRC as presented in Algorithm \ref{alg:multi}. It partitions the total $N$ iterations into $K$ stages, where stage $k$ contains $T_k$ iterations and $N=\sum_{k=1}^{K}T_k$. Within each stage $k$, SMRC performs $T_k$ updates using a fixed orthonormal basis $P_k$ and maintains the corresponding low-dimensional momentum $\rvm_{k,t}$. At the end of stage $k$, a new basis $P_{k+1}$ is computed from the dominant subspace of the stochastic gradient and the subsequent stage proceeds in the refreshed subspace.

The extension of the single-stage analysis to periodically refreshed bases introduces an additional challenge at stage boundaries. If each new stage is initialized with the terminal iterate of the preceding stage, the corresponding virtual sequence becomes \emph{inconsistent} across basis refreshes. To resolve this mismatch, we introduce a \emph{boundary correction} approach  that modifies the iterate at each stage transition to compensate for the offset. The corrected transition ensures the terminal virtual iterate of the stage coincides with the initial virtual iterate of the next.  The corresponding convergence guarantee of multi-stage SMRC is stated below.
\begin{thm}[Multi-stage SMRC]\label{thm:2} Let the total number of iterations be $N=\sum_{k=1}^{K}T_k$, where stage $k$ contains $T_k$ iterations. Define $Q_k=I-P_kP_k^\top$ and $\cc=(1-\beta)/\beta$.
Under  Assumptions~\ref{ass:smooth}, \ref{ass:lower:bound},
\ref{ass:var} and~\ref{ass:predictable}, for any orthonormal stage basis $P_k$ with $P_k^\top P_k=I_r$, by setting $\eta = \min\left\{\frac{\gamma}{2L},\frac{\gamma}{2L\cc}, \frac{1}{L\sqrt{N}}\right\}, \beta \in (0,1), \gamma \in (0,1]$, Algorithm~\ref{alg:multi} ensures
\begin{equation*}
    \frac{1}{N}\sum_{k=1}^K\sum_{t=1}^{T_k}\Ex{\big\|\nabla F(W_{k,t})\big\|_F} = O(N^{-1/4}).
\end{equation*}
\end{thm}

\begin{rem}\textnormal{ We have established an $O(N^{-1/4})$ convergence rate for multi-stage SMRC over $N$ iterations, which matches the convergence order of full-space momentum SGD \citep{Momentum:SGD}. Notably, Theorem \ref{thm:2} holds for any sequence of orthonormal bases $\{P_k\}_{k=1}^{K}$. In particular, each $P_k$ may be constructed from the dominant singular subspace of the stochastic gradient.}
\end{rem}

\section{Projected Muon with Residual Correction (PMRC)}
In this section, we extend the residual correction framework to Muon \citep{jordan2024muon}, a modern optimizer for LLM training. We first
present the single-stage method and establish its convergence guarantee and then extend the method and the guarantee to the multi-stage variant.
\begin{algorithm}[t]
\caption{Single-Stage PMRC}\label{alg:pmrc}
\textbf{Input}: initial iterate $W_1$, $P\in \R^{m\times r}$, $\eta$, $\gamma$, $\beta\in(0,1)$
\begin{algorithmic}[1]
\STATE Initialize momentum: $M_{0}=\textbf{0}$
\FOR{$t=1,\dots,T$}
  \STATE Query the stochastic gradient $G_t=g(W_t)$
  \STATE Calculate $M_t = (1-\beta)M_{t-1}+\beta P^\top G_t$ and $Z_t=\msign(M_t)$
  \STATE Update $W_{t+1}= W_t-\eta\big(PZ_t+\gamma (I-PP^\top) G_t\big)$
\ENDFOR
\end{algorithmic}
\end{algorithm}
\subsection{Single-Stage PMRC}
Muon \citep{jordan2024muon} has demonstrated strong performance in LLM training \citep{MuonScalable,KimiK2}, while its full-dimensional momentum incurs the same $O(mn)$ optimizer-state memory cost as momentum SGD. At larger model scales, this memory requirement becomes increasingly restrictive and motivates the development of memory-efficient variants of Muon \citep{pan2025gum}. A natural question is \emph{whether the momentum of Muon can be maintained in a low-dimensional subspace while preserving the convergence guarantee of full-space Muon}.

We incorporate the residual correction framework into Muon and develop PMRC, as outlined in Algorithm~\ref{alg:pmrc}. Specifically, PMRC projects the stochastic gradient $G_t$ onto the subspace, maintains the low-rank momentum $M_t = (1-\beta) M_{t-1} + \beta P^\top G_t \in\R^{r\times n}$ and applies the matrix sign to $M_t$. Then, PMRC incorporates the orthogonal residual gradient into the parameter update, that is, $W_{t+1} = W_t - \eta (P\operatorname{msign}(M_t) + \gamma (I-PP^\top)G_t)$, where $\gamma>0$ controls the residual correction. 

Notably, PMRC combines the Muon update with residual correction and the two update terms are characterized by \emph{different norm geometries.} The matrix-sign step corresponds to steepest descent under the spectral norm and is characterized by the nuclear norm \citep{bernstein2024oldoptimizernewnorm,li2025muon}, whereas the residual gradient correction is measured in the Frobenius norm. We analyze PMRC through a \emph{weighted momentum-tracking} argument for the bias-corrected projected momentum, which accounts for both momentum accumulation and the iterate variation induced by residual correction. We establish convergence under the following mixed stationarity measure.

\begin{thm}[Single-stage PMRC] \label{thm:3} Define $Q=I-PP^\top$. Under Assumptions~\ref{ass:smooth}, \ref{ass:lower:bound} and~\ref{ass:var}, for a matrix $P\in\R^{m\times r}$ with $P^\top P=I_r$ and by setting $\beta = \Theta(T^{-1/2}), \eta = \Theta(T^{-3/4}), \gamma = \Theta(T^{1/4})$ and $L\eta\gamma \leq 1$, Algorithm \ref{alg:pmrc} ensures
\begin{equation*}
\frac{1}{T}\sum_{t=1}^{T}\E\Big[\big\|P^\top\nabla F(W_t)\big\|_{*}+\big\|Q\nabla F(W_t)\big\|_F\Big] \le O(T^{-1/4}).
\end{equation*}
\end{thm}

\begin{rem}\textnormal{
Under standard stochastic assumptions, PMRC attains an $O(T^{-1/4})$ convergence rate under the mixed stationarity measure, matching the convergence order of full-space Muon \citep{li2025muon}. We establish convergence for a GaLore-style Muon variant that maintains its optimizer states entirely in a low-dimensional space.
}
\end{rem}
\begin{rem}
\textnormal{When $r=m$ and $P=I$, the mixed stationarity measure reduces to $\|\nabla F(W_t)\|_*$, which coincides with the standard stationarity measure for Muon. In this case, PMRC reduces to Muon and recovers the $O(T^{-1/4})$ convergence order of full-space Muon. Moreover, the mixed stationarity measure also upper bounds the Frobenius norm of the full gradient, i.e., $\|\nabla F(W_t)\|_F \le \|P^\top \nabla F(W_t)\|_* + \|(I-PP^\top)\nabla F(W_t)\|_F.$ Thus, convergence under the mixed stationarity measure also guarantees convergence of the full gradient in the Frobenius norm.}
\end{rem}

\begin{rem}\textnormal{
One may notice that the prescribed choice of $\gamma$ differs between SMRC and PMRC. Although $\gamma$ controls the weight of the residual correction relative to the projected update in both methods, the two algorithms are analyzed under different optimization geometries. In SMRC, both the projected and residual components are measured in the Frobenius norm, so  $\gamma\in(0,1]$ is sufficient for the convergence guarantee. In PMRC, the projected Muon component is measured in the nuclear norm, whereas the residual component is measured in the Frobenius norm. The choice $\gamma=\Theta(T^{1/4})$ corresponds to an effective step size $\eta\gamma=\Theta(T^{-1/2})$ for the residual component, the standard step size for SGD, while the Muon component retains $\eta=\Theta(T^{-3/4})$.
}\end{rem}

\begin{algorithm}[t]
\caption{Multi-Stage PMRC}\label{alg:pmrc-multi}
\textbf{Input}: initial iterate $W_{1,1}$, $P_1$, $\eta$, $\gamma$, $K$,  $T$, $\beta\in(0,1)$
\begin{algorithmic}[1]
\FOR{$k=1,\dots,K$}
  \STATE Set momentum $M_{k,0}=\textbf{0}$
  \FOR{$t=1,\dots,T$}
    \STATE Query the stochastic gradient $G_{k,t}=g(W_{k,t})$
    \STATE Calculate $M_{k,t}=(1-\beta)M_{k,t-1}+\beta P_k^\top G_{k,t}$ and  $Z_{k,t}=\msign(M_{k,t})$
    \STATE Update $W_{k,t+1}=W_{k,t}-\eta\big(P_k Z_{k,t}+\gamma(I-P_kP_k^\top)G_{k,t}\big)$
  \ENDFOR
  \IF{$k<K$}
    \STATE Set $W_{k+1,1}=W_{k,T+1}$ and update $P_{k+1}$
  \ENDIF
\ENDFOR
\end{algorithmic}
\end{algorithm}

\subsection{Multi-Stage PMRC}

Since the dominant gradient subspace may evolve during optimization, as discussed in Section~\ref{sec:smrc-multi}, we propose a multi-stage variant of PMRC, as presented in Algorithm~\ref{alg:pmrc-multi}.  At stage $k$, PMRC constructs an orthonormal basis $P_k$, initializes the momentum as $M_{k,0}=\mathbf{0}$, and performs $T$  updates with the basis $P_k$ fixed. At the end of stage $k$, the terminal iterate is used to initialize the subsequent stage, i.e., $W_{k+1,1}=W_{k,T+1}$. The projection basis is then refreshed to $P_{k+1}$, and the momentum is reinitialized in the corresponding projected coordinates. In this way, each stage retains the single-stage PMRC update with a fixed basis, while the projection subspace is periodically updated to track the evolving dominant gradient directions. We establish the convergence guarantee  below.

\begin{thm}[Multi-stage PMRC]\label{thm:4}
Define $Q_k=I-P_kP_k^\top$. Under Assumptions~\ref{ass:smooth}, \ref{ass:lower:bound},
\ref{ass:var} and~\ref{ass:predictable}, with stage lengths $T$, stage count $K=T$, total iterations $N=KT$ and by setting $\beta = \Theta(N^{-1/2}), \eta = \Theta(N^{-3/4})$, $\gamma=\Theta(N^{1/4})$ and $L\eta\gamma\le1$,
Algorithm~\ref{alg:pmrc-multi} ensures
\begin{equation*}
\frac{1}{N}\sum_{k=1}^{K}\sum_{t=1}^{T}
\E\Big[\big\|P_k^\top\nabla F(W_{k,t})\big\|_{*}+\big\|Q_k\nabla F(W_{k,t})\big\|_F\Big]
\le O(N^{-1/4}).
\end{equation*}
\end{thm}

\begin{rem}
\textnormal{
We establish an $O(N^{-1/4})$ convergence rate under the mixed stationarity measure, where $N$ is the total number of stochastic gradient evaluations. This rate is consistent with the single-stage guarantee in Theorem~\ref{thm:3} and the standard rate for full-space Muon \citep{li2025muon}. 
}
\end{rem}
 \begin{table*}[t!]
\centering
\small
\caption{Training and validation losses  during pretraining. \textbf{Bold} and \underline{underlined} values indicate the best and second-best results among memory-efficient methods, respectively.}
\label{tab:pretraining_loss}
\renewcommand{\arraystretch}{1.4}
\setlength{\tabcolsep}{8pt}
\begin{tabular}{ccccccc}
\hline
\textbf{Loss Type}
& \textbf{Memory-Efficient}
& \textbf{Optimizer}
& $\mathbf{60M}$
& $\mathbf{150M}$
& $\mathbf{350M}$
& $\mathbf{1.3B}$ \\
\hline
\multirow{7}{*}{\textbf{Training Loss}}
& $\textcolor{red}{\times}$
& AdamW
& $3.445$ & $3.156$ & $2.843$ & $2.551$ \\

& $\textcolor{red}{\times}$
& Muon
&$3.383$ & $3.086$ & $2.703$ & $2.414$ \\
\cline{2-7}

& $\textcolor{blue}{\checkmark}$
& GaLore
& $3.484$ & $3.180$ & $2.972$ & $2.842$ \\

& $\textcolor{blue}{\checkmark}$
& GoLore
& $3.516$ & $3.196$ & $3.003$ & $3.080$ \\

& $\textcolor{blue}{\checkmark}$
& Fira
& $3.601$ & $3.180$ & $\underline{2.852}$ & $2.657$ \\

& $\textcolor{blue}{\checkmark}$
& GUM
& $\underline{3.461}$ & $\underline{3.141}$ & $2.871$ & $\underline{2.609}$ \\

& $\textcolor{blue}{\checkmark}$
& \textbf{PMRC (Ours)}
& $\textbf{3.406}$ & $\textbf{3.117}$ & $\textbf{2.769}$ & $\mathbf{2.500}$ \\
\hline\hline

\multirow{7}{*}{\textbf{Validation Loss}}
& $\textcolor{red}{\times}$
& AdamW
& $3.421$ & $3.134$ & $2.892$ & $2.613$ \\

& $\textcolor{red}{\times}$
& Muon
&$3.352$ & $3.067$ & $2.746$ & $2.495$ \\
\cline{2-7}

& $\textcolor{blue}{\checkmark}$
& GaLore
& $3.462$ & $3.202$ & $2.980$ & $2.875$ \\

& $\textcolor{blue}{\checkmark}$
& GoLore
& $3.494$ & $3.204$ & $3.015$ & $3.110$ \\

& $\textcolor{blue}{\checkmark}$
& Fira
& $3.576$ & $3.161$ & $2.909$ & $2.730$ \\

& $\textcolor{blue}{\checkmark}$
& GUM
& $\underline{3.437}$ & $\underline{3.156}$ & $\underline{2.881}$ & $\underline{2.656}$ \\

& $\textcolor{blue}{\checkmark}$
& \textbf{PMRC (Ours)}
& $\textbf{3.380}$ & $\textbf{3.095}$ & $\textbf{2.786}$ & $\mathbf{2.570}$ \\
\hline
\end{tabular}
\end{table*}

\section{Empirical Study}
In this section, we assess the performance of the proposed methods through LLM pretraining and finetuning tasks. In experiments, we adopt a practical PMRC variant that applies a residual cap to improve training stability and refreshes the basis through an efficient approximation of SVD, as detailed in Appendix \ref{pratical}. Due to space limitations, we present only a subset of the results and defer the synthetic experiments and ablation studies to Appendix~\ref{sec:experiment} and Appendix~\ref{abla}, respectively.

\subsection{LLM pretraining}

We adopt the LLaMA architecture \citep{LLaMA2} and train models with $60$M, $150$M, $350$M and $1.3$B parameters. Their architectural configurations are summarized in Table~\ref{tab:model_architecture}. All models are pretrained on the C4 dataset \citep{C4}.

\textbf{Setup.} We train the $60$M and $150$M models on $2$ NVIDIA RTX PRO 6000 GPUs, the $350$M model on $4$ NVIDIA H20 GPUs, and the $1.3$B model on $8$ NVIDIA H20 GPUs. The corresponding per-device batch sizes are $256$, $256$, $160$, and $160$, yielding global batch sizes of $512$, $512$, $640$, and $1280$, respectively.  The sequence length is $512$ for the $60$M and $150$M models and $1024$ for the $350$M and $1.3$B models. For each model, the number of training steps approximately follows the Chinchilla scaling law \citep{hoffmann2022training}. For example, the $1.3$B model is trained for $20480$ steps, which corresponds to $20480\times1024\times1280\approx26.84$ billion training tokens, or approximately $20.6$ tokens per parameter. We compare our multi-stage PMRC with AdamW \citep{AdamW}, Muon \citep{jordan2024muon}, the AdamW-based GaLore \citep{GaLore}, GoLore \citep{GoLore} and Fira \citep{Fira}, and the Muon-based GUM \citep{pan2025gum}. For projection-based methods, we set the projection rank to $r=256$ and the refresh interval to $100$.
Since GUM randomly selects a subset of layers for full-rank updates, its memory cost depends on the number of full-rank layers $s$. For a fair comparison, we choose $s$ according to the memory budget of PMRC, setting $s=6$ and $r=128$ for the $60$M, $150$M, and $350$M models, and $s=4$ and $r=128$ for the $1.3$B model. The peak allocated memory for the $1.3$B model is reported in Table~\ref{tab:peak_gpu_memorypre}. The detailed setup, parameter settings and additional pretraining results are deferred to Appendix~\ref{exp-pt}.

\textbf{Results.} Table~\ref{tab:pretraining_loss} presents a comparison of the training and validation losses achieved by PMRC with those achieved by  AdamW, Muon,  GaLore, GoLore, Fira and GUM. As evidenced by the reported results, PMRC consistently achieves the lowest training and validation losses among memory-efficient methods. The improvement  supports the effectiveness of residual correction in practical LLM pretraining. PMRC also requires less optimizer-state memory than AdamW while achieving lower losses at every scale. We further evaluate the pretrained models on seven downstream benchmarks, with complete results reported in Appendix~\ref{exp-pt}, where PMRC achieves the highest average accuracy among memory-efficient methods.

\subsection{LLM supervised finetuning}
In this section, we conduct supervised finetuning (SFT) of LLMs to enhance their instruction-following and mathematical reasoning abilities. Following previous work \citep{pan2025gum,pan2025scalebio,yu2026stosignsgd}, we finetune Qwen3-8B \citep{yang2025qwen3}, LLaMA3.1-8B \citep{grattafiori2024llama} and OLMo-2-1124-7B \citep{olmo20242olmo2furious} on two common tasks. For a fair comparison, we set the projection rank to $r=512$ for all projection-based methods except GUM, for which we use $r=128$. For GUM, we adjust the number of selected full-rank layers according to the memory cost of PMRC, using $s=6$, $s=5$, and $s=5$ layers for Qwen3-8B, LLaMA3.1-8B, and  OLMo-2-1124-7B, respectively. Under these settings, the peak memory usage of GUM still exceeds that of PMRC.  We refresh the projection basis every $100$ optimizer steps for all memory-efficient optimizers. Our experiments are conducted on a single NVIDIA RTX PRO 6000 GPU and $4$ NVIDIA H20 GPUs.

\begin{table}[t]
\centering
\caption{Accuracy (\%) on IFEval and GSM8K benchmarks. \textbf{Bold} and \underline{underlined} values indicate the best and second-best results among memory-efficient methods, respectively.}
\label{tab:ifmain}
\small
\renewcommand{\arraystretch}{1.4}
\setlength{\tabcolsep}{5pt}
\setlength{\arrayrulewidth}{0.6pt}

\resizebox{\textwidth}{!}{%
\begin{tabular}{cccc|ccc}
\hline
& \multicolumn{3}{c}{\textbf{IFEval}}
& \multicolumn{3}{c}{\textbf{GSM8K}} \\
\cline{2-4}\cline{5-7}
\textbf{Optimizer}
& \textbf{Qwen3-8B}
& \textbf{LLaMA-3.1-8B}
& \textbf{OLMo-2-1124-7B}
& \textbf{Qwen3-8B}
& \textbf{LLaMA-3.1-8B}
& \textbf{OLMo-2-1124-7B} \\
\hline
AdamW
& $65.83$ & $50.12$ & $38.97$
& $90.14$ & $72.10$ & $77.63$  \\

Muon
& $62.11$ & $46.16$ & $36.45$
& $89.76$ & $72.78$  & $78.01$ \\
\hline
GaLore
& $66.67$ & $47.24$ & $\underline{38.13}$
& $ 89.16$ & $\underline{71.27}$ & $77.86$ \\

GoLore
& $\underline{68.35}$ & $\underline{47.96}$ & $35.73$
& $88.93$ & $70.66$ &  $77.10$ \\

GUM
& $62.47$ & $43.65$ & $36.21$
& $\underline{89.61}$ & $69.74$  & $\underline{78.32}$ \\

\textbf{PMRC (Ours)}
& $\textbf{69.06}$ & $\textbf{50.24}$ & $\mathbf{38.97}$
& $\textbf{90.30}$  & $\mathbf{72.48}$ & $\mathbf{78.62}$ \\
\hline
\end{tabular}%
}
\end{table}

\textbf{Setup.} For the instruction-following task, we follow \cite{pan2025gum} and train models on the GPT-4-LLM dataset, which contains approximately $52$K instruction-response examples synthesized by GPT-4. We report performance on IFEval \citep{zhou2023instruction}, a benchmark designed to measure how reliably a model satisfies explicit instructions and formatting constraints. For the mathematical reasoning task, we follow \citet{yu2026stosignsgd} and train models on the NuminaMath-CoT dataset \citep{numina_math_datasets}, which consists of $860$K problems from high school mathematics and international Olympiad competitions, and we randomly select $64$K of them for finetuning.  We evaluate the models on the GSM8K benchmark \citep{cobbe2021training}, which consists of grade school math word problems. For both tasks, we use a global batch size of $128$ and a maximum sequence length of $2048$, and train for one epoch. For evaluation, we allow up to $1280$ and $2048$ new tokens for IFEval and GSM8K, respectively.  We provide detailed training and evaluation configurations, along with peak allocated memory results, in Appendix~\ref{app:sft}.

\textbf{Results.} We report the instruction-level strict accuracy on the IFEval benchmark and the zero-shot accuracy on the GSM8K benchmark in Table~\ref{tab:ifmain}. On the IFEval benchmark, PMRC achieves the best performance among all projection-based methods across the three models, outperforming GUM by more than $2$ percentage points. Notably, on Qwen3-8B, PMRC also surpasses the full-space optimizers. On the GSM8K benchmark, PMRC achieves the best performance among all memory-efficient methods while remaining competitive with the full-space optimizers.
\section{Conclusion}

In this paper, we investigate memory-efficient subspace optimization and  adopt a residual correction framework. For momentum SGD, our proposed SMRC combines projected momentum with a gradient correction in the orthogonal complement and achieves an $O(T^{-1/4})$ convergence rate, which matches that of full-space momentum SGD.  We  generalize SMRC to a multi-stage variant with arbitrary basis refreshes and introduce a boundary correction that aligns consecutive virtual iterates, while retaining a convergence rate of the same order as the single-stage method.  Furthermore, we extend this framework to Muon and propose PMRC, which achieves an $O(T^{-1/4})$ rate under a mixed  stationarity measure,  while the multi-stage variant retains a convergence rate of the same order. The convergence rates of both PMRC variants are of the same order as that of full-space Muon. Both SMRC and PMRC reduce the optimizer-state memory from $O(mn)$ to $O((m+n)r)$. Extensive experimental results on LLM pretraining and supervised finetuning tasks demonstrate the practical effectiveness and efficiency of our proposed methods.

\subsection*{AI use statement}
In this work, we used generative AI tools for polishing, assistance with checking mathematical proofs,  creating or editing software code, suggesting experimental parameters, proof organization, and scripts for plotting results.  We reviewed all AI-assisted content. LLM-generated code was verified and tested for correctness by the authors. We take responsibility for the final content of this work, including text, claims or artifacts produced with the aid of generative AI. 





\bibliography{iclr2027_conference}
\bibliographystyle{iclr2027_conference}
\newpage
\appendix

\section{Additional Discussion on Related Work}\label{appendix:related}

\subsection{Additional Memory-Efficient Optimization Methods}\label{app:related1}

Several methods reduce memory cost through structured restrictions on the trainable parameters.  \cite{SLTrain} combine sparse and low-rank trainable components to increase the expressiveness of memory-efficient updates. LISA \citep{LISA} periodically resamples a subset of intermediate layers and updates  parameters within the selected layers.

Another line of work improves the construction and evolution of low-dimensional optimization subspaces. Flora \citep{FLORA} repeatedly resamples random low-rank projections, allowing updates from different subspaces to accumulate into higher-rank parameter changes. \cite{GRASS} construct sparse projection matrices from gradient row norms, which reduces the computational cost relative to GaLore. \cite{Online:SD} instead maintain the projection basis through online subspace estimation.  SARA \citep{SARA} samples singular directions according to their importance rather than deterministically retaining only the leading directions. \cite{RandomizedSubspace} formulate training over sampled low-dimensional parameter subspaces, reducing optimizer-state and activation memory.

\textbf{Comparison with \citet{SubspaceMomentum}.} \cite{SubspaceMomentum} combine projected momentum with an SGD update in the orthogonal complement. Their convergence guarantee is established for a fixed projection under stronger assumptions, with the residual update restricted to unit weight. It does not provide guarantees for subspace switching or the matrix-sign updates used by Muon. We address these limitations by analyzing weighted residual correction, establishing convergence for multi-stage SMRC with boundary correction, and developing a mixed-geometry analysis for PMRC. Moreover, the proof of Lemma~D.1 in \citet{SubspaceMomentum}
contains an error.
In our notation with $\gamma=1$ and $\rvm_0=\textbf{0}$,
the argument relies on the following inequalities
\begin{align*}
&-\eta\sum_{t=1}^{T} \left\langle P^\top\nabla F(W_t),\rvm_t\right\rangle\\
\leq&L\eta^2\sum_{t=1}^{T}\sum_{i=1}^{t-1} (1-\beta)^{t-i} \norm{P\rvm_i+QG_i}_F^2-\beta\eta\sum_{t=1}^{T}\sum_{i=1}^{t} (1-\beta)^{t-i} \left\langle P^\top\nabla F(W_i),P^\top G_i \right\rangle\\
=& L\eta^2\sum_{i=1}^{T-1}\norm{P\rvm_i+QG_i}_F^2\sum_{t=i+1}^{T}(1-\beta)^{t-i}-\beta\eta\sum_{i=1}^{T}\left\langle P^\top\nabla F(W_i),P^\top G_i\right\rangle\sum_{t=i}^{T}(1-\beta)^{t-i}\\
\leq & \frac{L\eta^2}{\beta}\sum_{t=1}^{T}\norm{P\rvm_t+QG_t}_F^2-\eta\sum_{t=1}^{T}\left\langle P^\top\nabla F(W_t),P^\top G_t\right\rangle.
\end{align*}
It omits the term $\eta\sum_{t=1}^{T}(1-\beta)^{T-t+1}\left\langle P^\top\nabla F(W_t),P^\top G_t\right\rangle.$ When $G_t=\nabla F(W_t)$, this quantity reduces to $ \eta\sum_{t=1}^{T}(1-\beta)^{T-t+1} \norm{P^\top\nabla F(W_t)}_F^2 \ge 0.$ Its removal can make the right-hand side smaller and is  not justified by the bound in the original paper. 

Other methods reduce memory by directly compressing the  optimizer states. Adam-mini \citep{AdamMini} replaces elementwise second-moment estimates with a scalar estimate shared within each Hessian-informed parameter block, while retaining the full first-moment estimate. APOLLO \citep{APOLLO} estimates channel-wise gradient scaling from low-rank first- and second-moment states constructed using  random projections.

\subsection{Low-rank Adaptation (LoRA)}

Low-Rank Adaptation (LoRA) \citep{LoRA} is an efficient finetuning method for LLMs. LoRA reduces the memory cost of finetuning by freezing the pretrained weights and training low-rank adapter matrices. Subsequent variants extend this approach through adaptive rank allocation across weight matrices \citep{AdaLoRA}, periodic merging of low-rank updates into the base weights \citep{ReLoRA}, and quantization of the frozen weights \citep{QLoRA}.

\subsection{Matrix-Aware and Muon-Based Optimizers}\label{app:muon-related}

Matrix-aware optimization exploits the structure of model parameters. Shampoo \citep{Shampoo} maintains factored preconditioners along individual dimensions. \cite{OneSidedShampoo} and \cite{an2025asgo} further construct adaptive preconditioners from structured matrix gradients. 

A separate line of work focuses on reducing the memory cost of Muon. SUMO \citep{refael2025sumo} orthogonalizes the momentum within a dynamically adapted low-dimensional subspace, but its convergence guarantee is established only for reversible layers. LiMuon \citep{LiMuon} combines momentum-based variance reduction with randomized SVD to compress the stored gradient estimate, reducing the memory required for optimizer states. \citet{QuantizedMuon} reduce the memory cost of Muon through blockwise 8-bit quantization, while \citet{su2026muonq} further compress the optimizer states to 4-bit precision.
\section{Proofs of the Theorems}
\subsection{Proof of Theorem \ref{thm:1}}
Throughout the proof, let $\{\mathcal{F}_t\}_{t=0}^T$ denote the
natural filtration, where $\mathcal{F}_{t-1}$ contains all randomness revealed before the stochastic gradient $G_t$ is queried. We define $Q=I-PP^\top$ and $\rvv_t=P\rvm_t$.  According to the momentum update of Algorithm~\ref{alg:single}, we have
\begin{equation}\label{eq:v-rec}
\rvv_t=(1-\beta)\rvv_{t-1}+\beta\,PP^\top G_t,\quad \rvv_0=\mathbf{0}.
\end{equation}
We denote $\cc=\frac{1-\beta}{\beta}$ and define the virtual iterate
\begin{equation}\label{eq:U-def}
U_t = W_t - \cc\eta\, \rvv_{t-1},
\end{equation}
so that $U_1=W_1$.
Then we define 
\begin{equation}\label{eq:C-def}
C = PP^\top + \gamma Q = \gamma I + (1-\gamma)PP^\top .
\end{equation}
Since $PP^\top$ and $Q$ are complementary orthogonal projectors, $C$ is symmetric with
eigenvalue $1$ of multiplicity $r$ and eigenvalue $\gamma$ of multiplicity $m-r$. In
particular, $0\prec\gamma I\preceq C\preceq I$ and $C^2\preceq C$ for $\gamma\in(0,1]$.

By substituting
the update $W_{t+1}=W_t-\eta(\rvv_t+\gamma QG_t)$ into \eqref{eq:U-def}, we can derive
\begin{equation}\label{eq:U-inter}
U_{t+1}=W_{t+1}-\cc\eta\,\rvv_t=W_t-\eta(1+\cc)\rvv_t-\eta\gamma QG_t .
\end{equation}
Since $1+\cc=1/\beta$, the recursion \eqref{eq:v-rec} implies
\begin{equation}\label{eq:v-id}
(1+\cc)\rvv_t=\beta^{-1}\rvv_t=\cc\,\rvv_{t-1}+PP^\top G_t .
\end{equation}
By substituting \eqref{eq:v-id} into \eqref{eq:U-inter} and using \eqref{eq:C-def}, we obtain
\begin{equation}\label{eq:U-rec}
U_{t+1}=(W_t-\cc\eta\,\rvv_{t-1})-\eta\big(PP^\top+\gamma Q\big)G_t=U_t-\eta\,CG_t .
\end{equation}

By using Assumption~\ref{ass:smooth} and~\eqref{eq:U-rec}, we have 
\begin{equation*}
    F(U_{t+1})\le F(U_t)-\eta\,\ip{\nabla F(U_t)}{CG_t}+\frac{L\eta^2}{2} \norm{CG_t}_F^2.
\end{equation*}
Furthermore, we have
\begin{equation}\label{eq:descent-cond}
\E\left[F(U_{t+1})\mid \mathcal{F}_{t-1}\right]\le F(U_t)-\eta\,\ip{\nabla F(U_t)}{C\,\nabla F(W_t)}+\frac{L\eta^2}{2}\,\E\left[\|C\,G_t\|_F^2\mid \mathcal{F}_{t-1}\right].
\end{equation}

We denote $\nabla F(U_t)=\nabla F(W_t)+\Delta_t$, where $\Delta_t=\nabla F(U_t)-\nabla F(W_t)$, and define
\begin{equation*}
    S_t=\ip{\nabla F(W_t)}{C\nabla F(W_t)}=\|PP^\top\nabla F(W_t)\|_F^2+\gamma\|Q\nabla F(W_t)\|_F^2.
\end{equation*}
Then we can derive
\begin{equation}\label{eq:3}
\ip{\nabla F(U_t)}{C\,\nabla F(W_t)}=S_t +\ip{\Delta_t}{C\,\nabla F(W_t)}.
\end{equation}

By using Assumption \ref{ass:smooth}, we have
\begin{equation*}
    \|\Delta_t\|_F\le L\|U_t-W_t\|_F=L\cc\eta\,\|\rvv_{t-1}\|_F. 
\end{equation*}
Moreover, since $\gamma\in(0,1]$, we have $\|C\|_{2}=1$. Therefore, we can derive
\begin{equation}\label{eq:1}
\big|\ip{\Delta_t}{C\nabla F(W_t)}\big| \le\|\Delta_t\|_F\,\|C\nabla F(W_t)\|_F\le
L\cc\eta\,\|\rvv_{t-1}\|_F\,\|C\nabla F(W_t)\|_F.
\end{equation}
By using $C^2\preceq C$, we have $\|C\nabla F(W_t)\|_F^2\le S_t$. Applying Young's inequality, we can further derive
\begin{equation}\label{eq:2}
\big|\ip{\Delta_t}{C\nabla F(W_t)}\big| \le\frac{1}{4}S_t+L^2\cc^2\eta^2\|\rvv_{t-1}\|_F^2.
\end{equation}
Combining \eqref{eq:3} and \eqref{eq:2}, we can obtain
\begin{equation}\label{eq:descent-lb}
\ip{\nabla F(U_t)}{C\,\nabla F(W_t)}
\ge\frac{3}{4}S_t-L^2\cc^2\eta^2\|\rvv_{t-1}\|_F^2.
\end{equation}

Moreover, since $C^2\preceq C\preceq I$, we have
$\|C\nabla F(W_t)\|_F^2\le S_t$ and
$\E[\|C(G_t-\nabla F(W_t))\|_F^2]\le\sigma^2$. Therefore, we obtain
\begin{equation}\label{eq:second-moment}
\begin{aligned}
\E\left[\|C\,G_t\|_F^2\mid \mathcal{F}_{t-1}\right]
&\le \|C\nabla F(W_t)\|_F^2   + \E\left[\|C(G_t-\nabla F(W_t))\|_F^2\mid \mathcal{F}_{t-1}\right] \\
&\le S_t+\sigma^2.
\end{aligned}
\end{equation}

Plugging~\eqref{eq:descent-lb} and~\eqref{eq:second-moment} into~\eqref{eq:descent-cond}, we can derive
\begin{equation*}
\E[F(U_{t+1})]\le\E[F(U_t)]-\eta\Big(\frac{3}{4}-\frac{L\eta}{2}\Big)\E[S_t]+L^2\cc^2\eta^3\Ex{\|\rvv_{t-1}\|_F^2}+\frac{L\eta^2}{2}\sigma^2.    
\end{equation*}

By setting $\eta\le \gamma/(2L)\le 1/(2L)$, we have $\frac{3}{4}-\frac{L\eta}{2}\ge \frac{1}{2}$ and can ensure
\begin{equation*}
\E[F(U_{t+1})]\le\E[F(U_t)]-\frac{\eta}{2}\E[S_t]+L^2\cc^2\eta^3\Ex{\|\rvv_{t-1}\|_F^2}+\frac{L\eta^2}{2}\sigma^2.
\end{equation*}
By summing up over $t=1,\dots,T$, we have
\begin{equation}\label{eq:telescoped}
\begin{split}
\frac{\eta}{2}\sum_{t=1}^{T}\Ex{S_t} \le & \E[F(W_1)]-\E[F(U_{T+1})] +L^2\cc^2\eta^3\sum_{t=1}^{T}\Ex{\|\rvv_{t-1}\|_F^2} +\frac{L\eta^2}{2}\,T\sigma^2.
\end{split}
\end{equation}

It remains to bound the term $\sum_{t=1}^{T}\Ex{\|\rvv_{t-1}\|_F^2}$. By using~\eqref{eq:v-rec}, we have 
\begin{align*}
\|\rvv_t\|_F^2
&=\big\|(1-\beta)\rvv_{t-1}+\beta\,PP^\top G_t\big\|_F^2\le(1-\beta)\|\rvv_{t-1}\|_F^2+\beta\|PP^\top G_t\|_F^2,
\end{align*}
which further implies $\beta\|\rvv_{t-1}\|_F^2 \le \|\rvv_{t-1}\|_F^2-\|\rvv_t\|_F^2+\beta\|PP^\top G_t\|_F^2$.
Summing over $t=1,\dots,T$ and using $\rvv_0=\textbf{0}$, we can derive $\sum_{t=1}^{T}\Ex{\|\rvv_{t-1}\|_F^2} \le \sum_{t=1}^{T}\Ex{\|PP^\top G_t\|_F^2}.$

According to Assumption \ref{ass:var} and $\|PP^\top\|_{2}\le 1$, we have
\begin{align*}
\sum_{t=1}^{T}\Ex{\|PP^\top G_t\|_F^2}\le\sum_{t=1}^{T}\Ex{\|PP^\top\nabla F(W_t)\|_F^2} + T\sigma^2\le
\sum_{t=1}^{T}\Ex{S_t} + T\sigma^2.    
\end{align*}

Finally, substituting these two inequalities into~\eqref{eq:telescoped}, we obtain
\begin{equation} \label{eqn:single:con1}
\Big(\frac{\eta}{2}-L^2\cc^2\eta^3\Big)\sum_{t=1}^{T}\E[S_t]
\le \E[F(W_1)]-\E[F(U_{T+1})]
+\Big(L^2\cc^2\eta^3+\frac{L\eta^2}{2}\Big)T\sigma^2.
\end{equation}

By setting $\eta\le \gamma/(2L\cc)\le 1/(2L\cc)$ when $\cc>0$, we have
$L^2\cc^2\eta^2\le \frac{1}{4}$ and  $\frac{\eta}{2}-L^2\cc^2\eta^3\ge \frac{\eta}{4}$. Therefore, \eqref{eqn:single:con1} implies
\begin{equation} \label{eqn:single:con2}
\begin{split}
\frac{\eta}{4}\sum_{t=1}^{T}\Ex{S_t}
\le & \E[F(W_1)]-\E[F(U_{T+1})]
+\Big(L^2\cc^2\eta^3+\frac{L\eta^2}{2}\Big)T\sigma^2\\
\leq & \E[F(W_1)]-F_*
+\Big(L^2\cc^2\eta^3+\frac{L\eta^2}{2}\Big)T\sigma^2,
\end{split}
\end{equation}
which  further implies
\begin{equation*}
    \frac{1}{T}\sum_{t=1}^{T}\E\Big[\big\|PP^\top\nabla F(W_t)\big\|_F^2+\gamma\big\|Q\nabla F(W_t)\big\|_F^2\Big] \le\frac{4\big(\E[F(W_1)]-F_*\big)}{\eta\,T}
+\big(2L\eta+4L^2\cc^2\eta^2\big)\sigma^2.
\end{equation*}
 
Since $S_t\ge\gamma\|\nabla F(W_t)\|_F^2$ and $\eta=\Theta(T^{-1/2})$
for fixed $\beta$ and $\gamma$, we obtain
\begin{equation*}
    \frac{1}{T}\sum_{t=1}^{T}\E\Big[\big\|\nabla F(W_t)\big\|_F^2\Big] \le O(T^{-1/2}).
\end{equation*}

Finally, applying  Jensen's inequality yields
\begin{equation*}
\begin{aligned}
\frac{1}{T}\sum_{t=1}^{T}\E\left[\|\nabla F(W_t)\|_F\right]
&\le
\left(
\frac{1}{T}\sum_{t=1}^{T}\E\left[\|\nabla F(W_t)\|_F^2\right]
\right)^{1/2} =O(T^{-1/4}).
\end{aligned}
\end{equation*}

\subsection{Proof of Theorem \ref{thm:2}}
For each stage $k\in\{1,\ldots,K\}$, let $\{\mathcal{F}_{k,t}\}_{t=0}^{T_k}$ denote the filtration, with $\mathcal{F}_{k,0}$ corresponding to the optimization history prior to stage $k$. By Assumption~\ref{ass:predictable}, $P_k$ is
$\mathcal{F}_{k,0}$-measurable and hence $\mathcal{F}_{k,t-1}$-measurable for every $t\in\{1,\ldots,T_k\}$.  Define $Q_k=I-P_kP_k^\top, \rvv_{k,t}=P_k \rvm_{k,t}$ and the virtual iterate $U_{k,t}=W_{k,t}-\cc\eta\,\rvv_{k,t-1}.$

Since the momentum is reset at the start of each stage, i.e.,
$\rvm_{k,0}=\mathbf{0}$, we have $\rvv_{k,0}=\mathbf{0}$ and  $U_{k,1}=W_{k,1}$. We define $S_{k,t}=\big\|P_kP_k^\top\nabla F(W_{k,t})\big\|_F^2
+\gamma\big\|Q_k\nabla F(W_{k,t})\big\|_F^2.$

Stage $k$ performs $T_k$ iterations of Algorithm~\ref{alg:single}, initialized at $W_{k,1}$ with $\rvm_{k,0}=\mathbf{0}$.  By setting $\eta \le \min\Big\{\frac{\gamma}{2L},\frac{\gamma}{2L\cc}, \frac{1}{L\sqrt{N}}\Big\}$, the single-stage
inequality~\eqref{eqn:single:con2} applies with $(W_t,G_t,U_t,T)$ replaced by $(W_{k,t},G_{k,t},U_{k,t},T_k)$. Taking expectations, we have
\begin{equation}\label{eq:stagewise-from-singlecon2}
\frac{\eta}{4}\sum_{t=1}^{T_k}\Ex{S_{k,t}}
\le\E\big[F(W_{k,1})\big]-\E\big[F(U_{k,T_k+1})\big]
+\Big(L^2\cc^2\eta^3+\frac{L\eta^2}{2}\Big)T_k\sigma^2 .
\end{equation}

By summing~\eqref{eq:stagewise-from-singlecon2} over $k=1,\dots,K$ and defining $N=\sum_{k=1}^K T_k$, we have
\begin{equation}\label{eq:stage-sum}
\frac{\eta}{4}\sum_{k=1}^{K}\sum_{t=1}^{T_k}\E[S_{k,t}]
\le\sum_{k=1}^{K}\Big(\E[F(W_{k,1})]-\E[F(U_{k,T_k+1})]\Big)
+\Big(L^2\cc^2\eta^3+\frac{L\eta^2}{2}\Big)N\sigma^2.
\end{equation}
Since we set $W_{k+1,1}=W_{k,T_k+1}-\cc\eta\,P_k\rvm_{k,T_k}$, we can ensure $W_{k+1,1}=U_{k,T_k+1}$.

Consequently, we further derive
\begin{equation}\label{eq:4}
    \sum_{k=1}^{K}\Big(\E[F(W_{k,1})]-\E[F(U_{k,T_k+1})]\Big)
=\E[F(W_{1,1})]-\E[F(U_{K,T_K+1})]
\le F(W_{1,1})-F_*,
\end{equation}
where the inequality uses Assumption~\ref{ass:lower:bound}.

By combining \eqref{eq:stage-sum} and \eqref{eq:4}, we have
\begin{equation*}
\frac{1}{N}\sum_{k=1}^{K}\sum_{t=1}^{T_k}\Ex{S_{k,t}} 
\le \frac{4\left(\Ex{F(W_{1,1})}-F_*\right)}{\eta N} + \left(2L\eta +4L^2\cc^2\eta^2\right)\sigma^2.
\end{equation*}

By further using $S_{k,t}\ge\gamma\|\nabla F(W_{k,t})\|_F^2$, we have
\begin{equation*}
\frac{1}{N}\sum_{k=1}^K\sum_{t=1}^{T_k}\Ex{\big\|\nabla F(W_{k,t})\big\|_F^2}
\le{}\frac{4\big(\E[F(W_{1,1})]-F_*\big)}{\gamma\,\eta\,N}+\Big(\frac{2L\eta}{\gamma}+\frac{4L^2\cc^2\eta^2}{\gamma}\Big)\sigma^2.
\end{equation*}

With fixed $\beta$ and $\gamma$, the choice
$\eta=\min\{1/(L\sqrt{N}),\gamma/(2L),\gamma/(2L\cc)\}$
satisfies $\eta=\Theta(N^{-1/2})$.
Finally, applying Jensen's inequality  yields
\begin{equation*}
\begin{aligned}
\frac{1}{N}\sum_{k=1}^{K}\sum_{t=1}^{T_k}
\E\left[\|\nabla F(W_{k,t})\|_F\right]
&\le
\left(
\frac{1}{N}\sum_{k=1}^{K}\sum_{t=1}^{T_k}
\E\left[\|\nabla F(W_{k,t})\|_F^2\right]
\right)^{1/2} =O(N^{-1/4}),
\end{aligned}
\end{equation*}
which completes the proof.

\subsection{Proof of Theorem \ref{thm:3}}

We first present several definitions. Let $\Delta=F(W_1)-F_*$ denote the initial optimality gap. For $X\in\R^{r\times n}$, $\msign(X)$ is equal to the solution of the problem $ \argmax_{\norm{Z}_2\le1}\ip{X}{Z}$. Hence we have
\begin{equation*}
    \norm{\msign(X)}_2\le1,
    \quad
    \norm{\msign(X)}_F^2\le r,
    \quad
    \ip{X}{\msign(X)}=\nuc{X}.
\end{equation*}
We define $A_t=P^\top\nabla F(W_t), R_t=Q\nabla F(W_t)$ and  $\xi_t=G_t-\nabla F(W_t).$ The corrected momentum and the corresponding tracking error are
\begin{equation*}
    \tilde M_t=\frac{M_t}{1-(1-\beta)^t}, \quad \tilde E_t=\tilde M_t-A_t, \quad  t\ge1.
\end{equation*}

We further define 
\begin{align*}
    \bar{A}=\frac1T\sum_{t=1}^T\E\left[\nuc{P^\top\nabla F(W_t)}\right],\qquad &\bar{R}=\frac1T\sum_{t=1}^T\E\left[\norm{Q\nabla F(W_t)}_F\right],\\
    \bar{S}=\frac1T\sum_{t=1}^T\E\left[\norm{Q\nabla F(W_t)}_F^2\right], \qquad &\overline{\mathcal E}=\frac1T\sum_{t=1}^T\E\left[\nuc{\tilde E_t}\right].
\end{align*}

Then we introduce several useful lemmas.
\begin{lem}[Lemma D.2 in \citet{yu2026sign}]\label{yu1}
    For any $X\in \R^{m\times n}$ and $a>0$, the following holds
    \begin{enumerate}
        \item $\ip{X}{\operatorname{msign}(X)}=\norm{X}_{*}$,
        \item $\operatorname{msign}(aX)=\operatorname{msign}(X)$.
    \end{enumerate}
\end{lem}
\begin{lem}[Lemma D.3 in \citet{yu2026sign}]\label{lem:muon:polar}
    For any $A,B\in\R^{r\times n}$ and any $Z_B=\msign(B)$,
\begin{equation*}
    \ip{A}{Z_B}\ge\nuc{A}-2\nuc{A-B}.
\end{equation*}
\end{lem}
\begin{lem}\label{lem:muon:ema}
Let $a=1-\beta$ and $\omega_{i,t}=\frac{\beta a^{t-i}}{1-a^t}, 1\le i\le t.$
Then we have
\begin{equation*}
    \tilde M_t=\sum_{i=1}^t\omega_{i,t}P^\top G_i, \quad
    \sum_{i=1}^t\omega_{i,t}=1, \quad v_t=\sum_{i=1}^t\omega_{i,t}^2
    =\frac{\beta}{2-\beta}\frac{1+a^t}{1-a^t}.
\end{equation*}
For $H\ge1$, we have $\frac1H\sum_{t=1}^H\sqrt{v_t}
    \le2\sqrt2\left(\sqrt\beta+H^{-1/2}\right).$
Moreover, for every $t\ge1$, we have
\begin{equation*}
    \sum_{i=1}^t\omega_{i,t}(t-i)\le\frac1\beta,
    \qquad
    \sum_{i=1}^t\omega_{i,t}\sqrt{t-i}\le\frac1{\sqrt\beta}.
\end{equation*}
\end{lem}
\begin{lem}\label{lem:muon:orth}
For each $Z\in\R^{r\times n}$ and $G\in\R^{m\times n}$, we have
\begin{equation*}
    \ip{PZ}{QG}=0, \qquad \norm{PZ+\gamma QG}_F^2
    =\norm{Z}_F^2+\gamma^2\norm{QG}_F^2.
\end{equation*}
In particular, if $Z=\msign(X)$ for some $X\in\R^{r\times n}$, then we have
\begin{equation*}
    \norm{PZ+\gamma QG}_F^2
    \le r+\gamma^2\norm{QG}_F^2.
\end{equation*}
\end{lem}

We start the proof by establishing a bound on $\bar{A}+\frac{\gamma}{2}\bar{S}$.
\begin{lem}\label{lem:muon:descent}
Under Assumptions \ref{ass:smooth}, \ref{ass:lower:bound} and \ref{ass:var}, if $L\eta\gamma\le1$, we have
\begin{equation*}
    \bar A+\frac{\gamma}{2}\bar S
    \le
    \frac{F(W_1)-F_*}{\eta T}
    +2\overline{\mathcal E}
    +\frac{L\eta r}{2}
    +\frac{L\eta\gamma^2\sigma^2}{2}.
\end{equation*}
Moreover, for each $t=1,\ldots,T$, we have
\begin{align*}
    \eta\Ex{\nuc{A_t}}
    +\frac{\eta\gamma}{2}\Ex{\norm{R_t}_F^2}
    \le{}&\E\bigl[F(W_t)-F(W_{t+1})\bigr]
    +2\eta\Ex{\nuc{\tilde E_t}}+\frac{L\eta^2r}{2}
    +\frac{L\eta^2\gamma^2\sigma^2}{2}.
\end{align*}
\end{lem}

It remains to derive a bound on $\overline{\mathcal E}$ and we present the following lemma.

\begin{lem}\label{lem:muon:tracking}
Under Assumptions \ref{ass:smooth}, \ref{ass:lower:bound} and \ref{ass:var}, we have 
\begin{align*}
    \overline{\mathcal E}
    \le{}&
    2\sqrt{2r}\,\sigma\left(\sqrt\beta+T^{-1/2}\right)
    +\frac{Lr\eta}{\beta}
    +\frac{L\sqrt r\,\eta\gamma\sigma}{\sqrt\beta}
    +\frac{L\sqrt r\,\eta\gamma}{\beta}\bar R.
\end{align*}
\end{lem}

By combining Lemma~\ref{lem:muon:descent} with Lemma \ref{lem:muon:tracking}, we have
\begin{align*}
    \bar A+\frac\gamma2\bar S
    \le{}&
    \frac{\Delta}{\eta T}
    +4\sqrt{2r}\,\sigma\left(\sqrt\beta+T^{-1/2}\right)
    +\frac{2Lr\eta}{\beta}
    +\frac{2L\sqrt r\,\eta\gamma\sigma}{\sqrt\beta}\\
    &+\frac{2L\sqrt r\,\eta\gamma}{\beta}\bar R
    +\frac{L\eta r}{2}
    +\frac{L\eta\gamma^2\sigma^2}{2}.
\end{align*}

Then we derive a bound on $\bar{R}$. By using Jensen's inequality and Young’s inequality, we have 
\begin{equation*}
    \frac{2L\sqrt r\,\eta\gamma}{\beta}\bar R
    \le
    \frac\gamma4\bar R^2
    +\frac{4L^2r\eta^2\gamma}{\beta^2}
    \le
    \frac\gamma4\bar S
    +\frac{4L^2r\eta^2\gamma}{\beta^2},
\end{equation*}
which further implies
\begin{equation}\label{eq:muon:single-general-refined}
\begin{aligned}
    \bar A+\frac\gamma4\bar S
    \le{}&
    \frac{\Delta}{\eta T}
    +4\sqrt{2r}\,\sigma\left(\sqrt\beta+T^{-1/2}\right)
    +\frac{2Lr\eta}{\beta}
    +\frac{2L\sqrt r\,\eta\gamma\sigma}{\sqrt\beta}\\
    &+\frac{L\eta r}{2}
    +\frac{L\eta\gamma^2\sigma^2}{2}
    +\frac{4L^2r\eta^2\gamma}{\beta^2}.
\end{aligned}
\end{equation}

We set $\beta = T^{-1/2}, \eta = \frac{1}{2L}T^{-3/4}, \gamma = T^{1/4}$, which satisfy $\gamma \eta L \leq 1$. According to \eqref{eq:muon:single-general-refined}, we have
\begin{equation*}
    \bar A+\frac\gamma4\bar S \leq \left[
        2L\Delta
        +\frac{9r}{4}
        +\frac{\sigma^2}{4}
        +(8\sqrt2+1)\sqrt r\,\sigma
    \right]T^{-1/4}.
\end{equation*}

To derive the guarantee for unsquared stationarity, we introduce a useful inequality.
For each $x\ge0$ and $\gamma>0$, we have
\begin{equation}\label{eq:muon:scalar-conversion}
    x\le\frac\gamma4x^2+\frac1\gamma.
\end{equation}
By using \eqref{eq:muon:scalar-conversion}, we can derive $\bar R\le\frac\gamma4\bar S+\frac1\gamma
    =\frac\gamma4\bar S+T^{-1/4}.$

Therefore, we can obtain
\begin{equation*}
    \frac{1}{T}\sum_{t=1}^T \Ex{\nuc{P^\top \nabla F(W_t)}+\norm{Q\nabla F(W_t)}_F}\leq\left[
        2L\Delta
        +\frac{9r}{4}
        +\frac{\sigma^2}{4}
        +(8\sqrt2+1)\sqrt r\,\sigma
        +1
    \right]T^{-1/4}. 
\end{equation*}

Moreover, by using $\norm{X}_F \leq \norm{P^\top X}_{*} + \norm{QX}_F$, we can derive 
\begin{equation*}
    \frac{1}{T}\sum_{t=1}^T \Ex{\norm{\nabla F(W_t)}_F}\leq\left[
        2L\Delta
        +\frac{9r}{4}
        +\frac{\sigma^2}{4}
        +(8\sqrt2+1)\sqrt r\,\sigma
        +1
    \right]T^{-1/4}. 
\end{equation*}

\subsection{Proof of Theorem \ref{thm:4}}
First, we present several definitions. Let $\Delta=F(W_{1,1})-F_*$ denote the initial optimality gap. Consider $K$ stages of length $T$. The total number of gradient evaluations is $N=KT$. At the start of stage $k$, the algorithm selects a predictable orthonormal basis $P_k\in\R^{m\times r}$ and sets $Q_k=I-P_kP_k^\top$, $M_{k,0} = \mathbf{0}$. We define 
\begin{equation*}
A_{k,t}=P_k^\top\nabla F(W_{k,t}),\quad    R_{k,t}=Q_k\nabla F(W_{k,t}),\quad \tilde M_{k,t}=\frac{M_{k,t}}{1-(1-\beta)^t},\quad \tilde{E}_{k,t}=\tilde M_{k,t}-A_{k,t}.
\end{equation*}
\begin{align*}
    \bar A&=\frac1N\sum_{k=1}^K\sum_{t=1}^T\Ex{\nuc{A_{k,t}}},\qquad
    \bar R=\frac1N\sum_{k=1}^K\sum_{t=1}^T\Ex{\norm{R_{k,t}}_F},\\
    \bar S&=\frac1N\sum_{k=1}^K\sum_{t=1}^T\Ex{\norm{R_{k,t}}_F^2},\qquad
    \overline{\mathcal E}=\frac1N\sum_{k=1}^K\sum_{t=1}^T\Ex{\nuc{\tilde E_{k,t}}}.
\end{align*}

As in the previous proof, we first establish the guarantee for multi-stage descent. We can apply Lemma \ref{lem:muon:descent} and replace $(P,Q,W_t,G_t,A_t,R_t,\tilde E_t)$ by $(P_k,Q_k,W_{k,t},G_{k,t},A_{k,t},R_{k,t},\tilde E_{k,t}).$ Thus, we can derive
\begin{align*}
    \eta\Ex{\nuc{A_{k,t}}}
    +\frac{\eta\gamma}{2}\Ex{\norm{R_{k,t}}_F^2}
    \le{}&\E\bigl[F(W_{k,t})-F(W_{k,t+1})\bigr]
    +2\eta\Ex{\nuc{\tilde E_{k,t}}}\\
    &+\frac{L\eta^2r}{2}
    +\frac{L\eta^2\gamma^2\sigma^2}{2}.
\end{align*}

Summing over $t=1,\ldots,T$ and $k=1,\ldots,K$ and using $ \sum_{k=1}^K\sum_{t=1}^T
      \E\bigl[F(W_{k,t})-F(W_{k,t+1})\bigr]
    =F(W_{1,1})-\E[F(W_{K,T+1})]
    \le F(W_{1,1})-F_*$, we have
\begin{equation}\label{eq:muon:ms-descent}
    \bar A+\frac\gamma2\bar S
    \le
    \frac{F(W_{1,1})-F_*}{\eta N}
    +2\overline{\mathcal E}
    +\frac{L\eta r}{2}
    +\frac{L\eta\gamma^2\sigma^2}{2}.
\end{equation}
Then we bound the term $\overline{\mathcal E}$. By applying Lemma~\ref{lem:muon:tracking}, we can derive
\begin{align*}
    \frac1T\sum_{t=1}^T\E\left[\nuc{\tilde E_{k,t}}\right]
    \le{}&
    2\sqrt{2r}\,\sigma\left(\sqrt\beta+T^{-1/2}\right)
    +\frac{Lr\eta}{\beta}
    +\frac{L\sqrt r\,\eta\gamma\sigma}{\sqrt\beta}\\
    &+\frac{L\sqrt r\,\eta\gamma}{\beta}
      \frac{1}{T}\sum_{t=1}^T\E\left[\norm{R_{k,t}}_F\right].
\end{align*}

Since $N = KT$, we can derive
\begin{equation}\label{eq:muon:ms-tracking}
\begin{aligned}
    \overline{\mathcal E}
    \le{}&
    2\sqrt{2r}\,\sigma\left(\sqrt\beta+T^{-1/2}\right)
    +\frac{Lr\eta}{\beta}
    +\frac{L\sqrt r\,\eta\gamma\sigma}{\sqrt\beta}
    +\frac{L\sqrt r\,\eta\gamma}{\beta}\bar R.
\end{aligned}
\end{equation}

Combining \eqref{eq:muon:ms-descent} and \eqref{eq:muon:ms-tracking} and using $\frac{2L\sqrt r\eta\gamma}{\beta}\bar R\le\frac{\gamma}{4}\bar S+\frac{4L^2r\eta^2\gamma}{\beta^2}$, we have
\begin{align*}
    \bar A+\frac\gamma4\bar S
    \le{}&
    \frac{\Delta}{\eta N}
    +4\sqrt{2r}\,\sigma\left(\sqrt\beta+T^{-1/2}\right)
    +\frac{2Lr\eta}{\beta}
    +\frac{2L\sqrt r\,\eta\gamma\sigma}{\sqrt\beta}\\
    &+\frac{L\eta r}{2}
    +\frac{L\eta\gamma^2\sigma^2}{2}
    +\frac{4L^2r\eta^2\gamma}{\beta^2}.
\end{align*}

By setting $\eta = \frac{1}{2L}N^{-3/4}, \beta = N^{-1/2}, \gamma = N^{1/4}, T = K = \sqrt{N}$, we have 
\begin{align*}
    &\frac1N\sum_{k=1}^K\sum_{t=1}^T\E\left[
        \nuc{P_k^\top\nabla F(W_{k,t})}
        +\frac\gamma4\norm{Q_k\nabla F(W_{k,t})}_F^2
    \right]\\
    &\quad\le
    \left[
        2L\Delta
        +\frac{9r}{4}
        +\frac{\sigma^2}{4}
        +(8\sqrt2+1)\sqrt r\,\sigma
    \right]N^{-1/4}.
\end{align*}

By using \eqref{eq:muon:scalar-conversion}, we can obtain
\begin{align*}
&\frac1N\sum_{k=1}^K\sum_{t=1}^T\E\left[\nuc{P_k^\top\nabla F(W_{k,t})}
+\norm{Q_k\nabla F(W_{k,t})}_F\right]\\
    &\quad\le\left[2L\Delta+\frac{9r}{4}+\frac{\sigma^2}{4}+(8\sqrt2+1)\sqrt r\,\sigma+1\right]N^{-1/4}.
\end{align*}

According to $\norm{X}_F \leq \norm{P_k^\top X}_{*} + \norm{Q_kX}_F$, we also have 
\begin{align*}
&\frac1N\sum_{k=1}^K\sum_{t=1}^T\Ex{\norm{\nabla F(W_{k,t})}_F}\le\left[2L\Delta+\frac{9r}{4}+\frac{\sigma^2}{4}+(8\sqrt2+1)\sqrt r\,\sigma+1\right]N^{-1/4}.
\end{align*}

\section{Proofs of Supporting Lemmas}

\subsection{Proof of Lemma \ref{lem:muon:ema}}

First, it is not hard to verify that $M_t=\beta\sum_{i=1}^t a^{t-i}P^\top G_i.$ By summing $\omega_{i,t}$ from $i=1$ to $t$, we have
\begin{align*}
    \sum_{i=1}^t\omega_{i,t}
    =\frac{\beta}{1-a^t}\sum_{j=0}^{t-1}a^j
    =\frac{\beta}{1-a^t}\frac{1-a^t}{1-a}=1.
\end{align*}
The weights satisfy
\begin{align*}
    \sum_{i=1}^t\omega_{i,t}^2 &=\frac{\beta^2}{(1-a^t)^2}\sum_{j=0}^{t-1}a^{2j} =\frac{\beta^2(1-a^{2t})}{(1-a^2)(1-a^t)^2} =\frac{\beta}{2-\beta}\frac{1+a^t}{1-a^t},
\end{align*}
where the identity is due to $1-a^2=\beta(2-\beta)$ and $1-a^{2t}=(1-a^t)(1+a^t)$.

By using $(1-\beta)^t\le(1+\beta t)^{-1}$, we have $1-a^t\ge\frac{\beta t}{1+\beta t}.$ Furthermore, by using  $1+a^t\le2$ and $2-\beta\ge1$, we can derive $v_t\le2\beta+\frac{2}{t},\sqrt{v_t}\le\sqrt{2\beta}+\sqrt{\frac2t}.$

Furthermore, we can derive
\begin{align*}
    \sqrt{2\beta}+\frac{\sqrt2}{H}\sum_{t=1}^Ht^{-1/2} \le\sqrt{2\beta}+2\sqrt2\,H^{-1/2}\le2\sqrt2\left(\sqrt\beta+H^{-1/2}\right),
\end{align*}
where the first inequality is due to $\sum_{t=1}^Ht^{-1/2}\le2\sqrt H$. Let $\ell=t-i$. According to the definition of $\omega_{i,t}$, we obtain
\begin{equation*}
    \sum_{i=1}^t \omega_{i,t}(t-i)
=
\frac{\beta}{1-a^t}
\sum_{\ell=0}^{t-1}\ell a^\ell.
\end{equation*}
Moreover, we can derive
\begin{equation*}
    \sum_{\ell=0}^{t-1}\ell a^\ell
=
\frac{a-ta^t+(t-1)a^{t+1}}{(1-a)^2}
\end{equation*}
and we have 
\begin{align*}
\sum_{i=1}^t \omega_{i,t}(t-i)
&=
\frac{a-ta^t+(t-1)a^{t+1}}
     {\beta(1-a^t)} =
\frac{a}{\beta}
-
\frac{ta^t}{1-a^t} \le
\frac{a}{\beta}
\le
\frac{1}{\beta}.
\end{align*}
Moreover, $\sum_{i=1}^t\omega_{i,t}=1$. By using Jensen's inequality, we obtain
\begin{align*}
\sum_{i=1}^t\omega_{i,t}\sqrt{t-i}
&\le
\left(
\sum_{i=1}^t\omega_{i,t}(t-i)
\right)^{1/2} \le
\sqrt{\frac{a}{\beta}}
\le
\frac{1}{\sqrt{\beta}}.
\end{align*}

\subsection{Proof of Lemma \ref{lem:muon:orth}}
Since $P^\top Q=P^\top-P^\top PP^\top=0$, we can derive $\ip{PZ}{QG}=\tr(Z^\top P^\top QG)=0.$
We also have $\norm{PZ}_F^2=\tr(Z^\top P^\top PZ)=\norm{Z}_F^2.$ Therefore, we can derive
\begin{align*}
    \norm{PZ+\gamma QG}_F^2=\norm{Z}_F^2+\gamma^2\norm{QG}_F^2 + 2\gamma  \ip{PZ}{QG}
    =\norm{Z}_F^2+\gamma^2\norm{QG}_F^2.
\end{align*}

If $Z=\msign(X)$, then we have $\norm{Z}_F^2\le r$ and 
\begin{align*}
    \norm{PZ+\gamma QG}_F^2=\norm{Z}_F^2+\gamma^2\norm{QG}_F^2 + 2\gamma  \ip{PZ}{QG}
    \leq r+\gamma^2\norm{QG}_F^2.
\end{align*}

\subsection{Proof of Lemma~\ref{lem:muon:descent}}

According to Assumption \ref{ass:smooth}, we can derive
\begin{align*}
    F(W_{t+1})
    \le{}&F(W_t)
    -\eta\ip{\nabla F(W_t)}{PZ_t}
    -\eta\gamma\ip{\nabla F(W_t)}{QG_t}
    +\frac{L\eta^2}{2}\norm{PZ_t+\gamma QG_t}_F^2.
\end{align*}
According to  Lemma~\ref{lem:muon:polar}, we further have 
\begin{equation*}
    \ip{\nabla F(W_t)}{PZ_t}
    =\ip{A_t}{\msign(\tilde{M}_t)}
    \ge\nuc{A_t}-2\nuc{\tilde E_t}.
\end{equation*}

Since $Q^\top  = Q$ and $Q^2 = Q$, we can derive $\E\!\left[\ip{\nabla F(W_t)}{QG_t}\right]=\Ex{\ip{R_t}{Q\nabla F(W_t)}}=\Ex{\norm{R_t}_F^2}.$

By using Lemma~\ref{lem:muon:orth}, we have $ \Ex{\norm{PZ_t+\gamma QG_t}_F^2} \le r+\gamma^2\Ex{\norm{QG_t}_F^2}.$ Therefore, by using $\E\!\left[\norm{QG_t}_F^2\right]\le\Ex{\norm{R_t}_F^2}+\sigma^2,$ we can derive

\begin{align*}
\E[F(W_{t+1})]
\le{}&
\E[F(W_t)]
-\eta\E[\nuc{A_t}]
+2\eta\E\left[\nuc{\tilde E_t}\right] -\eta\gamma\left(1-\frac{L\eta\gamma}{2}\right)
\E\left[\norm{R_t}_F^2\right]
+\frac{L\eta^2r}{2}
+\frac{L\eta^2\gamma^2\sigma^2}{2}.
\end{align*}

Summing up from $t=1$ to $T$ and using $L\eta \gamma \leq 1$, we conclude the proof.

\subsection{Proof of Lemma \ref{lem:muon:tracking}}

By using Lemma \ref{lem:muon:ema}, $P^\top G_i=A_i+P^\top\xi_i$ and $\sum_{i=1}^t\omega_{i,t}=1$, we have 
\begin{equation*}
\tilde E_t = \underbrace{ \sum_{i=1}^{t}\omega_{i,t}P^\top\xi_i }_{\mathrm{Term-1}} + \underbrace{ \sum_{i=1}^{t}\omega_{i,t}(A_i-A_t)}_{\mathrm{Term-2}}.
\end{equation*}
Then we bound the two terms separately. First we have 
\begin{equation*}
    \Ex{\norm{\sum_{i=1}^t\omega_{i,t}P^\top\xi_i}_F^2}\leq \sum_{i=1}^t\omega_{i,t}^2\E\left[\norm{P^\top\xi_i}_F^2\right]\le\sigma^2v_t.
\end{equation*}

By using $\norm{X}_*\leq \sqrt{r}\norm{X}_{F}, \forall X\in \R^{r\times n}, r\leq n$, we can derive 
\begin{equation*}
    \Ex{\norm{\sum_{i=1}^t\omega_{i,t}P^\top\xi_i}_*}\leq \sigma\sqrt{rv_t}.
\end{equation*}

By using Lemma \ref{lem:muon:ema}, we obtain 
\begin{equation*}
\frac1T\sum_{t=1}^T\E\left[\nuc{\sum_{i=1}^t\omega_{i,t}P^\top\xi_i}\right]
    \le2\sqrt{2r}\sigma\left(\sqrt\beta+T^{-1/2}\right).
\end{equation*}

Under Assumption \ref{ass:smooth}, we have $\nuc{A_i-A_t}\le\sqrt r\,L\norm{W_i-W_t}_F.$ According to the update in Algorithm \ref{alg:pmrc}, for $1\leq i\leq t$ we can ensure 
\begin{align*}
    \norm{W_i-W_t}_F
    \le{}&\eta\sqrt r\,(t-i)
    +\eta\gamma\sum_{s=i}^{t-1}\norm{R_s}_F
    +\eta\gamma\norm{\sum_{s=i}^{t-1}Q\xi_s}_F.
\end{align*}

To derive a bound on $\mathrm{Term-2}$, we first have 
\begin{align*}
    \norm{\omega_{i,t}(A_i-A_t)}_*&\leq L \omega_{i,t}\eta r(t-i)
    +L\omega_{i,t}\sqrt{r}\eta\gamma\sum_{s=i}^{t-1}\norm{R_s}_F
    +L \omega_{i,t}\sqrt{r}\eta\gamma\norm{\sum_{s=i}^{t-1}Q\xi_s}_F.
\end{align*}

Then we bound these terms one by one.
\begin{align*}
\frac1T\sum_{t=1}^T\sum_{i=1}^t\omega_{i,t}
      \sum_{s=i}^{t-1}\Ex{\norm{R_s}_F} &=\frac1T\sum_{t=1}^T\sum_{s=1}^{t-1}
      \left(\sum_{i=1}^s\omega_{i,t}\right)\Ex{\norm{R_s}_F} \\&\le\frac1T\sum_{t=1}^T\sum_{s=1}^{t-1}
      a^{t-s}\Ex{\norm{R_s}_F}\\
&=\frac1T\sum_{s=1}^{T-1}\Ex{\norm{R_s}_F}
      \sum_{j=1}^{T-s}a^j\\
&\le\frac{a}{\beta}\frac1T\sum_{s=1}^T\Ex{\norm{R_s}_F}
      \le\frac{\bar R}{\beta}.
\end{align*}

We also have 
\begin{equation*}
    \Ex{\norm{\sum_{s=i}^{t-1}Q\xi_s}_F^2}
    =\sum_{s=i}^{t-1}\Ex{\norm{Q\xi_s}_F^2}
    \le(t-i)\sigma^2,
\end{equation*}
which further implies 
\begin{equation*}
    \Ex{\norm{\sum_{s=i}^{t-1}Q\xi_s}_F}
    \le\sigma\sqrt{t-i}.
\end{equation*}

By using Lemma \ref{lem:muon:ema}, we have
\begin{align*}
    \frac1T\sum_{t=1}^T\sum_{i=1}^t\omega_{i,t}
    \E\left[\norm{\sum_{s=i}^{t-1}Q\xi_s}_F\right]
    &\le\frac{\sigma}{T}\sum_{t=1}^T\sum_{i=1}^t
       \omega_{i,t}\sqrt{t-i} \le\frac{\sigma}{\sqrt\beta}.
\end{align*}
Finally, we can obtain
\begin{equation*}
    \Ex{\frac{1}{T}\sum_{t=1}^T \norm{\sum_{i=1}^t\omega_{i,t}(A_i-A_t)}_*} \le
    \frac{Lr\eta}{\beta}
    +\frac{L\sqrt r\,\eta\gamma}{\beta}\bar R
    +\frac{L\sqrt r\,\eta\gamma\sigma}{\sqrt\beta}.
\end{equation*}

\subsection{Proof of Lemma~\ref{lem:1}}
Since  $\rvm_t = (1-\beta)\rvm_{t-1}+\beta P^\top G_t$, we can verify 
\begin{equation*}
    P \rvm_t = (1-\beta)P \rvm_{t-1}+\beta P P^\top G_t.
\end{equation*}
Then, we have
\begin{align*}
&\|P P^\top \nabla F(W_t) - P \rvm_t\|_F^2 = \|P P^\top \nabla F(W_t) - (1-\beta)P \rvm_{t-1}-\beta P P^\top G_t\|_F^2 \\
=& \left\|(1-\beta) \big( P P^\top \nabla F(W_t) - P \rvm_{t-1} \big)+\beta P P^\top \big( \nabla F(W_t) - G_t \big) \right\|_F^2.
\end{align*}
By taking expectations, we have
\begin{equation} \label{eqn:lem1:1}
\begin{split}
&\E \left[\|P P^\top \nabla F(W_t) - P \rvm_t\|_F^2 \right] \\ 
= & (1-\beta)^2 \E \left[ \left\| P P^\top \nabla F(W_t) - P \rvm_{t-1} \right\|_F^2  \right]+ \beta^2 \E \left[ \left\| P P^\top \big( \nabla F(W_t) - G_t \big) \right\|_F^2 \right]\\
\leq & (1-\beta)^2 (1+\beta) \E \left[ \left\|  P P^\top \nabla F(W_{t-1}) - P \rvm_{t-1}  \right\|_F^2  \right]\\
& + (1-\beta)^2 \left(1+\frac{1}{\beta}  \right) \E \left[ \left\|P P^\top \nabla F(W_t) - P P^\top \nabla F(W_{t-1}) \right\|_F^2  \right] + \beta^2 \sigma^2\\
\leq & (1-\beta) \E \left[ \left\|  P P^\top \nabla F(W_{t-1}) - P \rvm_{t-1} \right\|_F^2  \right] + \frac{1}{\beta} \E \left[ \left\|P P^\top \nabla F(W_t) - P P^\top \nabla F(W_{t-1}) \right\|_F^2  \right] + \beta^2 \sigma^2,
\end{split}
\end{equation}
 where the first inequality is due to the weighted Young inequality $\|x+y\|_F^2 \le (1+\theta)\|x\|_F^2 + (1+1/\theta)\|y\|_F^2$. To bound the second term on the right-hand side, we use Assumption \ref{ass:smooth} and obtain
\begin{align*}
    &\left\|P P^\top \nabla F(W_t) - P P^\top \nabla F(W_{t-1}) \right\|_F^2
\leq \left\|\nabla F(W_t) -\nabla F(W_{t-1}) \right\|_F^2\\
\leq &L^2  \left\|W_t - W_{t-1}\right\|_F^2
=L^2  \left\|\eta P\rvm_{t-1}+ \eta\gamma (I-PP^\top) G_{t-1} \right\|_F^2 \\
= & \eta^2 L^2  \left\| P\rvm_{t-1} \right\|_F^2
+ \eta^2 \gamma^2 L^2 \left\|  (I-PP^\top) G_{t-1} \right\|_F^2,
\end{align*}
where the last term is the undesirable contribution from the SGD correction, which causes this direct decomposition argument to break down. To make this explicit, we substitute the above bound into~\eqref{eqn:lem1:1} and obtain
\begin{align*}
&\E \left[\|P P^\top \nabla F(W_t) - P \rvm_t\|_F^2 \right] \\ 
\leq &(1-\beta) \E \left[ \left\|  P P^\top \nabla F(W_{t-1}) - P \rvm_{t-1} \right\|_F^2  \right]  + \frac{\eta^2 L^2}{\beta}   \E \left[\left\| P\rvm_{t-1} \right\|_F^2 \right]\\
&+ \frac{\eta^2 \gamma^2 L^2}{\beta}  \E \left[ \left\|  (I-PP^\top) G_{t-1} \right\|_F^2 \right] + \beta^2 \sigma^2.
\end{align*}

Summing over $t=2,\ldots,T+1$, we have
\begin{align*}
&\E \left[\sum_{t=2}^{T+1}\|P P^\top \nabla F(W_t) - P \rvm_t\|_F^2 \right] \\ 
\leq &(1-\beta)  \E \left[ \sum_{t=2}^{T+1}\left\|  P P^\top \nabla F(W_{t-1}) - P \rvm_{t-1} \right\|_F^2  \right]  + \frac{\eta^2 L^2}{\beta} \E \left[ \sum_{t=2}^{T+1}  \left\| P\rvm_{t-1} \right\|_F^2 \right]\\
&+ \frac{\eta^2 \gamma^2 L^2}{\beta}  \E \left[ \sum_{t=2}^{T+1} \left\|  (I-PP^\top) G_{t-1} \right\|_F^2\right] + \beta^2 \sigma^2 T\\
= &(1-\beta) \E \left[ \sum_{t=1}^{T} \left\|  P P^\top \nabla F(W_{t}) - P \rvm_{t} \right\|_F^2  \right]  + \frac{\eta^2 L^2}{\beta}  \E \left[\sum_{t=1}^{T}  \left\| P\rvm_{t} \right\|_F^2 \right]\\
&+ \frac{\eta^2 \gamma^2 L^2}{\beta}  \E \left[ \sum_{t=1}^{T} \left\|  (I-PP^\top) G_{t} \right\|_F^2\right] + \beta^2 \sigma^2 T.
\end{align*}
Rearranging the above inequality, we obtain \eqref{eqn:lem1:2}.

\section{Limitations of a Direct Decomposition Analysis} \label{sec:Orthogonal}
This section explains why a seemingly simpler orthogonal-decomposition proof does not yield the desired rate. According to Assumption \ref{ass:smooth}, we have
\begin{align*}
F(W_{t+1}) \leq &F(W_t) + \langle  \nabla F(W_t), W_{t+1}-W_t \rangle + \frac{L}{2}  \|W_{t+1}-W_t\|_F^2 \\
=&F(W_t) - \langle  \nabla F(W_t), \eta P \rvm_t+ \eta\gamma (I-PP^\top) G_t \rangle + \frac{L}{2}  \| \eta P \rvm_t+ \eta\gamma (I-PP^\top) G_t\|_F^2.
\end{align*}

Since $P \rvm_t$ and $(I-PP^\top) G_t$ are orthogonal and the quadratic term expands without a cross term, we can derive
\begin{equation}\label{eqn:thm1:1}
\begin{split}
F(W_{t+1}) \leq &  F(W_t)    \underbrace{-\eta \langle  \nabla F(W_t),  P \rvm_t \rangle + \frac{\eta^2 L}{2}  \| P \rvm_t\|_F^2}_{\mathrm{Term-1}(t)} \\
 &  \underbrace{-\eta\gamma \langle  \nabla F(W_t), (I-PP^\top) G_t \rangle + \frac{\eta^2\gamma^2 L}{2}  \|   (I-PP^\top) G_t\|_F^2}_{\mathrm{Term-2}(t)}.
\end{split}
\end{equation}
Here, $\mathrm{Term-1}(t)$ corresponds to the momentum-SGD update in $\SPN(P)$, whereas $\mathrm{Term-2}(t)$ corresponds to the residual SGD correction in $\SPN(P)^\perp$. A natural approach is to analyze $\mathrm{Term-1}(t)$ using a standard momentum-tracking argument and $\mathrm{Term-2}(t)$ using a standard SGD descent analysis, and then combine the resulting bounds. We show below that this direct decomposition fails to establish the desired convergence rate.

To control $\mathrm{Term-1}(t)$, we follow existing work \citep{ICML:2023:Jiang} and have
\begin{align*}
\mathrm{Term-1}(t) =& -\eta \langle  \nabla F(W_t),  P \rvm_t \rangle + \frac{\eta}{2} \|P P^\top \nabla F(W_t)\|_F^2 + \frac{\eta}{2} \|P \rvm_t \|_F^2 \\
&  - \frac{\eta}{2} \|P P^\top \nabla F(W_t)\|_F^2 - \frac{\eta}{2} \|P \rvm_t \|_F^2 + \frac{\eta^2 L}{2}  \| P \rvm_t\|_F^2 \\
\leq & \frac{\eta}{2}  \|P P^\top \nabla F(W_t) - P \rvm_t\|_F^2 - \frac{\eta}{2} \|P P^\top \nabla F(W_t)\|_F^2  - \frac{\eta}{4} \|P \rvm_t \|_F^2, 
\end{align*}
where the inequality is due to $\eta \leq 1/(2L)$. Summing the above inequality over $t=1,\ldots, T$, we have
\begin{equation}\label{eqn:thm1:2}
\begin{split}
\sum_{t=1}^{T} \mathrm{Term-1}(t)
\leq & \frac{\eta}{2} \sum_{t=1}^{T}
\|P P^\top \nabla F(W_t) - P \rvm_t\|_F^2 - \frac{\eta}{2} \sum_{t=1}^{T}
\|P P^\top \nabla F(W_t)\|_F^2
- \frac{\eta}{4} \sum_{t=1}^{T} \|P \rvm_t \|_F^2 .
\end{split}
\end{equation}
It remains to bound the first term, which quantifies the tracking error between $P\rvm_t$ and $PP^\top \nabla F(W_t)$. To this end, we have the following lemma.
\begin{lem} \label{lem:1} For the momentum estimator in Algorithm \ref{alg:single}, we have
\begin{equation} \label{eqn:lem1:2}
\begin{split}
&\E\left[ \sum_{t=1}^{T}\|P P^\top \nabla F(W_t) - P \rvm_t\|_F^2 \right] \\
 \leq & \frac{\E\left[ \|P P^\top \nabla F(W_1) - P \rvm_1\|_F^2 - \|P P^\top \nabla F(W_{T+1}) - P \rvm_{T+1}\|_F^2 \right]}{\beta} + \frac{\eta^2 L^2}{\beta^2}  \E \left[\sum_{t=1}^{T}  \left\| P\rvm_{t} \right\|_F^2 \right]\\
&+ \frac{\eta^2 \gamma^2 L^2}{\beta^2}  \E \left[ \sum_{t=1}^{T} \left\|  (I-PP^\top) G_{t} \right\|_F^2\right]+ \beta \sigma^2 T .
\end{split}
\end{equation}
\end{lem}

Combining \eqref{eqn:lem1:2} and \eqref{eqn:thm1:2}, and using  $\eta \leq \beta/(2L)$, we have
\begin{equation}\label{eqn:thm1:3}
\begin{split}
 \E\left[\sum_{t=1}^{T} \mathrm{Term-1}(t) \right]\leq & \frac{\eta}{2 \beta} \left(\E\left[ \|P P^\top \nabla F(W_1) - P \rvm_1\|_F^2 - \|P P^\top \nabla F(W_{T+1}) - P \rvm_{T+1}\|_F^2 \right]\right) \\
     & + \frac{\eta \beta \sigma^2T }{2}
     - \frac{\eta}{2}  \E\left[\sum_{t=1}^{T} \|P P^\top \nabla F(W_t)\|_F^2\right] + \frac{\eta^2 \gamma^2 L}{4 \beta}
     \E \left[ \sum_{t=1}^{T} \left\|  (I-PP^\top) G_{t} \right\|_F^2\right].
\end{split}
\end{equation}
The last term in~\eqref{eqn:thm1:3} is where this route fails. Lemma~\ref{lem:1} must control the change of $PP^\top\nabla F(W_t)$ between consecutive iterates. Smoothness bounds this change by $L\|W_t-W_{t-1}\|_F$ and the step contains the residual correction $\eta\gamma(I-PP^\top)G_{t-1}$, so the residual enters the tracking-error bound with coefficient $\eta^2\gamma^2L/(4\beta)$. The residual descent in $\sum_t \mathrm{Term-2}(t)$ has coefficient $\eta\gamma/2$, so the two differ by a factor $\eta\gamma L/(2\beta)$. The remaining terms in~\eqref{eqn:thm1:3} force $\beta=\Theta(1/\sqrt{T})$ and $\eta=\Theta(1/\sqrt{T})$ and for constant $\gamma$ this factor is then $\Theta(1)$. 

Next, we bound $\mathrm{Term-2}(t)$ based on a standard analysis:
\begin{align*}
&\E \left[\mathrm{Term-2}(t)\right]\\
=& \E\left[-\eta\gamma \langle  \nabla F(W_t), (I-PP^\top) G_t \rangle + \frac{\eta^2\gamma^2 L}{2}  \|   (I-PP^\top) G_t\|_F^2\right] \\
=&-\eta\gamma \E \left[\| (I-PP^\top) \nabla F(W_t) \|_F^2\right] +  \frac{\eta^2\gamma^2 L}{2} \E \left[ \|   (I-PP^\top) ( G_t -\nabla F(W_t))+ (I-PP^\top)\nabla F(W_t) \|_F^2\right]\\
\leq & -\eta\gamma \E \left[\| (I-PP^\top) \nabla F(W_t) \|_F^2\right] +  \frac{\eta^2\gamma^2 L}{2} \E \left[ \| (I-PP^\top)\nabla F(W_t) \|_F^2\right]+ \frac{\eta^2\gamma^2 L \sigma^2}{2} \\
\leq & -\frac{\eta\gamma}{2} \E \left[\| (I-PP^\top) \nabla F(W_t) \|_F^2\right]+ \frac{\eta^2\gamma^2 L \sigma^2}{2},
\end{align*}
where the last inequality is due to $\eta \gamma \leq 1/L$. Summing the above inequality over $t=1,\ldots, T$, we have
\begin{equation}\label{eqn:thm1:4}
\begin{split}
\E \left[\sum_{t=1}^{T} \mathrm{Term-2}(t)\right] \leq -\frac{\eta\gamma}{2} \E \left[\sum_{t=1}^{T} \| (I-PP^\top) \nabla F(W_t) \|_F^2\right]+ \frac{\eta^2\gamma^2 L \sigma^2 T}{2} .
\end{split}
\end{equation}

Summing (\ref{eqn:thm1:1}) over all rounds, taking expectations, and setting $\gamma\leq1$, we have
\begin{align*}
\E\left[ F(W_{T+1}) -F(W_1) \right] &\leq 
-\frac{\eta\gamma}{2}
\E \left[\sum_{t=1}^{T} \|  \nabla F(W_t) \|_F^2\right]+  \frac{\eta^2\gamma^2 L \sigma^2 T}{2}+\frac{\eta \beta \sigma^2T }{2}  \\
&+\frac{\eta}{2 \beta}\E\Big[
\|P P^\top \nabla F(W_1) - P \rvm_1\|_F^2
- \|P P^\top \nabla F(W_{T+1}) - P \rvm_{T+1}\|_F^2 \Big]\\
 & + \frac{\eta^2 \gamma^2 L}{4 \beta}
\E \left[ \sum_{t=1}^{T} \left\|  (I-PP^\top) G_{t} \right\|_F^2\right],
\end{align*}
where the inequality is due to \eqref{eqn:thm1:3} and \eqref{eqn:thm1:4}.
Finally, we have
\begin{equation}\label{eqn:thm1:5}
\begin{split}
&\frac{1}{T}\E \left[\sum_{t=1}^{T} \|  \nabla F(W_t) \|_F^2\right] \\
\leq & \frac{2}{\eta \gamma T} \E\left[F(W_1) -  F(W_{T+1}) \right] \\
&+ \frac{1}{\beta \gamma T} \left(\E\left[ \|P P^\top \nabla F(W_1) - P \rvm_1\|_F^2 - \|P P^\top \nabla F(W_{T+1}) - P \rvm_{T+1}\|_F^2 \right]\right) \\
& + \left(\eta\gamma L +\frac{ \beta  }{  \gamma}  \right)  \sigma^2 + \frac{\eta \gamma L}{2\beta T}
\E \left[ \sum_{t=1}^{T} \left\|  (I-PP^\top) G_{t} \right\|_F^2\right].
\end{split}
\end{equation}

Without the last term in~\eqref{eqn:thm1:5}, the choices $\gamma=\Theta(1)$ and $\eta,\beta=\Theta(T^{-1/2})$ would yield an $O(T^{-1/2})$ bound on the average squared gradient norm.  However, under this scaling the coefficient $\eta\gamma L/(2\beta)$ remains $\Theta(1)$.
Although the true-gradient component of the residual second moment can be absorbed into the descent term, bounding its noise component by $\sigma^2$ leaves a variance contribution that does not vanish with $T$. Consequently, this tracking bound does not establish the desired $O(T^{-1/2})$ rate under stochastic gradients.

\section{Experiments}\label{exp}

In this section, we  assess the performance of the proposed methods via numerical experiments.

\subsection{Synthetic Experiments}\label{sec:experiment}
We conduct two synthetic experiments  and compare our methods with Momentum SGD, GaLore~\citep{GaLore} and GoLore~\citep{GoLore}. 
\subsubsection{Effect of Residual Correction}

\textbf{Setup.} We conduct this experiment using a stochastic quadratic problem adapted from the counterexample of \citet{GoLore}:
\begin{equation}\label{sexp-1}
F_1(W)=\frac{1}{2}\|AW\|_F^2+\langle B,W\rangle,\qquad
G_t=\nabla F_1(W_t)+\xi_t,    
\end{equation}
where $W\in\R^{m\times n}$ and $\xi_t$ is gradient noise with $\Ex{\xi_t}=0$. We set $m=30, n=8$ and the projection rank $r=2$. Because the number of rows exceeds the number of columns, all low-rank methods use a right projection basis $P_t\in\mathbb{R}^{8\times 2}$. For GaLore, GoLore and multi-stage SMRC, we set the refresh interval to $50$. We train all methods for $400$ iterations and grid search the parameters $\eta\in
\{0.05,0.1,0.2,0.5,0.8, 0.9\}$ and $\beta\in
\{0.1,0.3,0.5,0.7,0.9\}$. We set $\gamma = 0.8$ for SMRC.
We set
\begin{equation*}
U_s
=
\begin{bmatrix}
I_2\\
\mathbf{0}_{28\times2}
\end{bmatrix}
\in\mathbb{R}^{30\times2},
\quad
U_n
=
\begin{bmatrix}\mathbf{0}_{28\times2}\\
I_2
\end{bmatrix}
\in\mathbb{R}^{30\times2},
\end{equation*}
which satisfy $U_s^\top U_s=I_2, U_n^\top U_n=I_2, U_s^\top U_n=\mathbf{0}.$  We further define
\begin{equation*}
H=
\begin{bmatrix}
1&1&1&1&1&1&1&1\\
1&-1&1&-1&1&-1&1&-1
\end{bmatrix}.
\end{equation*}
The quadratic objective is specified by $A=U_s^\top, B=\frac{1}{2}U_sH,$ where $A\in\mathbb{R}^{2\times30}$ has orthonormal rows. The parameter matrix is initialized as $W_0
= U_s\operatorname{diag}(1.8,0.9)H.$ 
We  define $V_s=\frac{1}{\sqrt{8}}H^\top$  and compute a complete QR factorization of $V_s$ and form
$V_n\in\mathbb{R}^{8\times2}$ from the third and fourth columns of the orthogonal matrix. We have $V_n^\top V_n=I_2, V_n^\top V_s=\mathbf{0}.$ At iteration $t$, the stochastic gradient is $G_t=\nabla F_1(W_t)+\xi_t,$ where $\xi_t = 4s_tU_nV_n^\top, s_t\overset{\mathrm{i.i.d.}}{\sim} \operatorname{Rademacher}(\{-1,+1\}).$ 

\begin{figure*}[t!]
\begin{center}
\includegraphics[draft=false,width=0.4\textwidth]{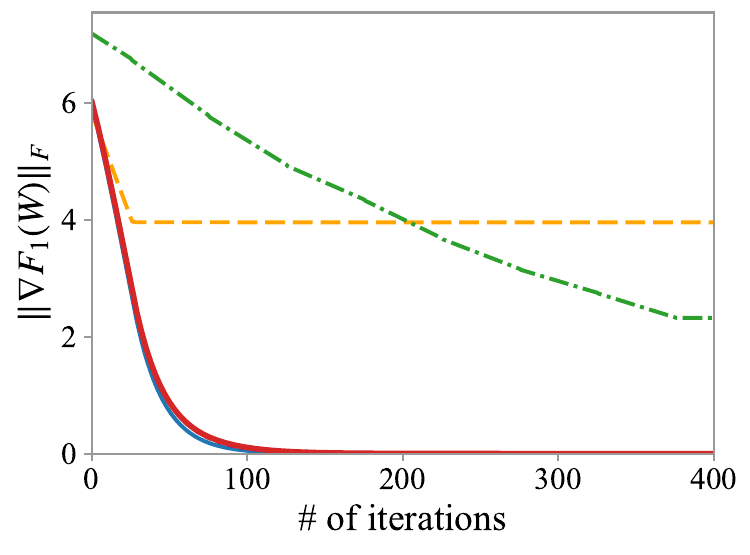}
\includegraphics[draft=false,width=0.4\textwidth]{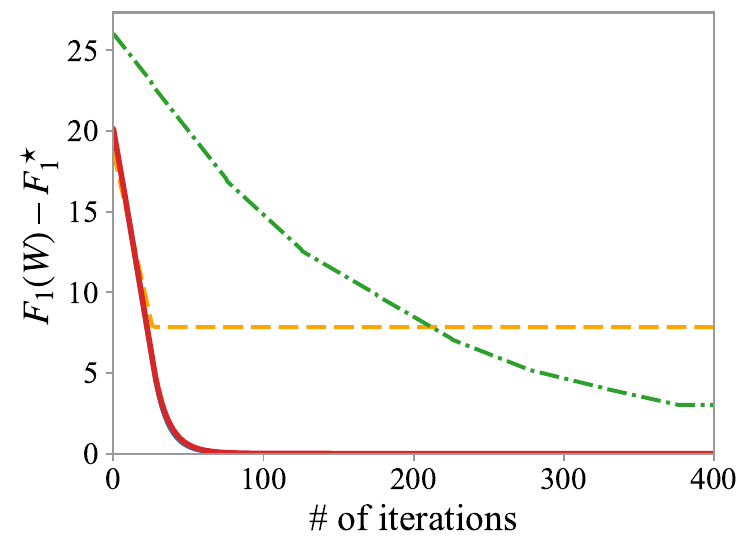}
\subfigure{
\includegraphics[draft=false,width=0.9\textwidth]{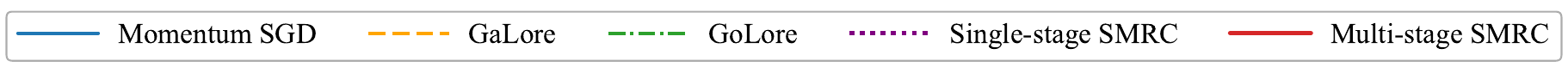}
}
\caption{Effect of residual correction.}
\label{fig:residual-correction}
\end{center}
\end{figure*}

\textbf{Results.} As shown in Figure~\ref{fig:residual-correction}, GaLore fails to converge when its selected subspace becomes dominated by gradient noise, leaving informative gradient directions outside the subspace. By incorporating the discarded gradient components through residual correction, SMRC restores convergence and achieves performance comparable to that of Momentum SGD.

\subsubsection{Benefit of Multi-Stage Subspace Refreshing}
\textbf{Setup.} We consider the following problem
\begin{equation}\label{sexp-2}
F_2(W)=
\frac{\lambda_1}{2}\|U_1^\top W\|_F^2+
\frac{\lambda_2}{2}\|U_2^\top W\|_F^2+
\frac{\mu}{2}\|U_\perp^\top W\|_F^2,
\qquad
G_t=\nabla F_2(W_t)+\xi_t,
\end{equation}
where $W\in\mathbb{R}^{30\times32}$. Let $e_i\in\mathbb{R}^{30}$
be the $i$-th standard unit vector. We set
$U_1=[e_1,e_2]$, $U_2=[e_3,e_4]$,
$U_\perp=[e_5,\ldots,e_{30}]$, 
$(\lambda_1,\lambda_2,\mu)=(4,0.02,0.01)$ and  $W_0=1.2U_1H_1+50U_2H_2$, where
\begin{equation*}
H_1=
\begin{bmatrix}
\mathbf{1}_8^\top&\mathbf{0}_8^\top&\mathbf{0}_8^\top&\mathbf{0}_8^\top\\
\mathbf{0}_8^\top&\mathbf{1}_8^\top&\mathbf{0}_8^\top&\mathbf{0}_8^\top
\end{bmatrix},
\qquad
H_2=
\begin{bmatrix}
\mathbf{0}_8^\top&\mathbf{0}_8^\top&\mathbf{1}_8^\top&\mathbf{0}_8^\top\\
\mathbf{0}_8^\top&\mathbf{0}_8^\top&\mathbf{0}_8^\top&\mathbf{1}_8^\top
\end{bmatrix}.
\end{equation*}
We use  Gaussian noise $\xi_t=0.01Z_t$, where
$(Z_t)_{ij}\overset{\mathrm{i.i.d.}}{\sim}\mathcal{N}(0,1)$.
We set the projection rank to $r=2$ and train for $300$ iterations. Single-stage SMRC keeps the subspace fixed, whereas multi-stage SMRC refreshes it every $20$ iterations. At each stage transition, multi-stage SMRC explicitly applies the boundary correction in Algorithm~\ref{alg:multi}. For all methods, we search $\eta,\beta\in\{0.02,0.05,0.1,0.3,0.5\}$. For SMRC variants, we additionally search over $\gamma\in\{0.02,0.05,0.1,0.3,0.5,0.7\}$.

\begin{figure*}[t!]
\begin{center}
\includegraphics[draft=false,width=0.4\textwidth]{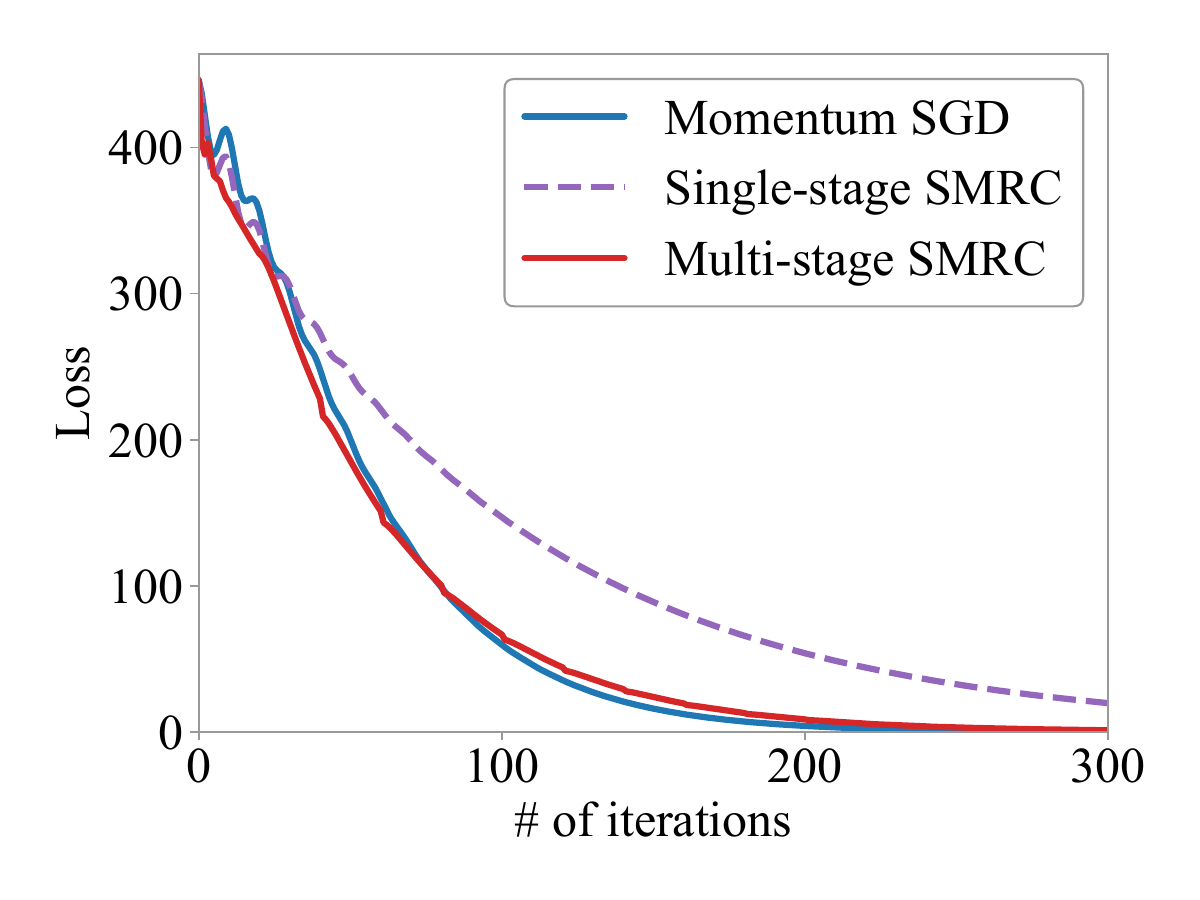}
\caption{Results of multi-stage refreshing.}
\label{fig:multistage-refresh}
\end{center}
\end{figure*}

\textbf{Results.} Figure~\ref{fig:multistage-refresh} shows that single-stage SMRC continues to make progress through residual correction, but its convergence slows as the dominant gradient directions evolve during optimization. In contrast, multi-stage SMRC periodically refreshes the projection basis to track the evolving dominant directions, resulting in faster convergence and performance comparable to that of full-space Momentum SGD.

\subsection{Language Model Experiments}
To verify the empirical effectiveness of the proposed algorithm in practice, we compare our PMRC with previous methods in LLM pretraining and finetuning settings.

\subsubsection{A practical implementation of PMRC}\label{pratical}

We present a practical variant of Algorithm~\ref{alg:pmrc-multi} with the following two modifications.

First, although the residual correction is explicitly scaled by $\gamma$, a single global coefficient cannot consistently align its magnitude with the projected Muon direction across parameter blocks and optimization steps. In particular, during early training, the initial subspace may leave a relatively large fraction of the gradient in the orthogonal complement, causing $\gamma\lVert(I-P_kP_k^\top)G_{k,t}\rVert_F$ to substantially exceed $\lVert P_kZ_{k,t}\rVert_F$. We observed that this scale imbalance coincides with pronounced loss fluctuations during the early stage of training. We introduce a residual cap that bounds the residual correction with respect to the projected Muon direction, while preserving the original coefficient $\gamma$ whenever the residual update is already below the prescribed threshold.

Define $R_{k,t}=(I-P_kP_k^\top)G_{k,t}$. We modify the update rule as follows:
\begin{equation}\label{finalupdate}
W_{k,t+1}=W_{k,t}-0.2\sqrt{\max\{m,n\}}\,\eta\left(P_kZ_{k,t}+\widehat{\gamma}_{k,t}R_{k,t}\right)-\eta\lambda W_{k,t},
\end{equation}
where
\begin{equation*}
\widehat{\gamma}_{k,t}=\min\left\{\gamma,\rho\sqrt{\frac{\lVert P_kZ_{k,t}\rVert_F^2}{\max\{\lVert R_{k,t}\rVert_F^2,\epsilon\}}}\right\}.
\end{equation*}
Following the standard practical implementation of Muon, we retain the shape-dependent learning-rate scaling factor $0.2\sqrt{\max\{m,n\}}$ and weight decay with coefficient $\lambda$. These two components are inherited from \cite{MuonScalable}. We set $\rho=4$ and $\epsilon=10^{-30}$ in our experiments.

Second, periodic SVD computations in GaLore introduce time overhead when the projection basis is refreshed, particularly for high-dimensional matrices. To improve computational efficiency, we replace the exact SVD with a warm-started subspace approximation adapted from ARO \citep{gong2026aronewlensmatrix}. Specifically, given the matrix $G\in\R^{m\times n}$ and the previous basis $P^{\mathrm{prev}}\in\R^{m\times r}$, we update the basis by
\begin{equation*}
P=\operatorname{QR}\!\left( G\,f_{\mathrm{RN}}\!\left((P^{\mathrm{prev}})^\top G\right)^\top \right),
\end{equation*}
where $f_{\mathrm{RN}}(\cdot)$ performs row normalization and $\operatorname{QR}(\cdot)$ returns an orthonormal basis. The corresponding singular values are estimated from the row norms of $P^\top G$. In our implementation, the QR operation is computed with SCQR \citep{fukaya2020shifted}. 
\begin{table}[t]
\centering
\small
\caption{Per-matrix projector latency on the LLaMA-350M matrix shapes on a single NVIDIA RTX PRO 6000 GPU.}
\label{tab:refresh_time}
\renewcommand{\arraystretch}{1.3}
\setlength{\tabcolsep}{6pt}
\begin{tabular}{lcc}
\hline
\textbf{Matrix Shape}
& \textbf{Exact SVD (ms)}
& \textbf{Ours (ms)}\\
\hline
$1024 \times 1024$
& $58.01$
& $3.21$\\
$2736 \times 1024$
& $57.75$
& $2.98$ \\
$1024 \times 2736$
& $57.36$
& $2.98$ \\
\hline
\end{tabular}
\end{table}
To quantify the practical benefit, we measure the per-matrix latency of a basis refresh on the LLaMA-350M shapes. As shown in Table~\ref{tab:refresh_time}, the approximate refresh consistently takes around $3$ milliseconds, whereas the exact SVD used by GaLore requires nearly $58$ milliseconds.

\begin{table*}[t]
\centering
\small
\caption{Architecture configurations of the LLaMA-style models.}
\label{tab:model_architecture}
\renewcommand{\arraystretch}{1.3}
\setlength{\tabcolsep}{6pt}
\begin{tabular}{lcccccc}
\hline
\textbf{Model Size}
& \textbf{Hidden Size}
& \textbf{Layers}
& \textbf{Heads}
& \textbf{FFN Size}
& \textbf{Vocabulary Size}
& \textbf{Max Sequence Length} \\[2pt]
\hline
$60\mathrm{M}$
& $512$
& $8$
& $8$
& $1376$
& $32000$
& $1024$ \\
$150\mathrm{M}$
& $768$
& $14$
& $12$
& $2048$
& $32000$
& $2048$ \\
$350\mathrm{M}$
& $1024$
& $24$
& $16$
& $2736$
& $32000$
& $2048$ \\
$1.3\mathrm{B}$
& $2048$
& $24$
& $32$
& $5461$
& $32000$
& $1024$ \\
\hline
\end{tabular}
\end{table*}

\begin{table*}[t]
\centering
\small
\caption{Training configurations of the LLaMA-style models.}
\label{tab:training_configurations}

\renewcommand{\arraystretch}{1.3}
\setlength{\tabcolsep}{12pt}

\begin{tabular}{lcccc}
\hline
\textbf{Hyperparameter}
& $60\mathrm{M}$
& $150\mathrm{M}$
& $350\mathrm{M}$
& $1.3\mathrm{B}$ \\
\hline
Global Batch Size
& $512$ &$512$ &$640$ &$1280$ \\

Device Micro-batch Size
& $256$&$256$ &$160$ &$160$ \\

Sequence Length
&$512$ &$512$ &$1024$ &$1024$ \\

Tokens per Step
& $262144$ & $262144$ & $655360$ & $1310720$ \\

Training Steps
& $4431$ & $9900$ & $11300$ & $20480$ \\

Total Training Tokens
& $1.162$B & $2.595$B & $7.406$B & $26.84$B \\

Learning-rate Schedule
& cosine &cosine &cosine & cosine\\

Gradient Clipping
& $1.0$ & $1.0$ &$1.0$ &  $1.0$\\
\hline
\end{tabular}
\end{table*}

\begin{table}[t]
    \centering
    \caption{Peak allocated GPU memory (GiB) during $1.3$B pretraining.}
    \label{tab:peak_gpu_memorypre}
    \renewcommand{\arraystretch}{1.2}
    \resizebox{0.7\linewidth}{!}{%
        \begin{tabular}{cccccccc}
            \hline
         \textbf{Model} & \textbf{AdamW} & \textbf{Muon} & \textbf{GaLore} & \textbf{GoLore} & \textbf{Fira} & \textbf{GUM} & \textbf{PMRC} \\
            \hline
        \textbf{1.3B} &  $57.77$ & $55.52$ & $54.00$  & $54.00$  & $54.00$ & $53.85$  & $\textbf{53.44}$ \\
            \hline
        \end{tabular}%
    }
\end{table}

\begin{table}[h]
\centering
\caption{Hyperparameter settings for different optimizers and model sizes.}
\label{tab:optimizer-hparams}
\renewcommand{\arraystretch}{1.15}

\resizebox{\linewidth}{!}{%
\begin{tabular}{ccccccccc}
\hline
\textbf{Model}
& \textbf{AdamW}
& \textbf{Muon}
& \textbf{GaLore}
& \textbf{GoLore}
& \textbf{Fira}
& \textbf{GUM}
& \multicolumn{2}{c}{\textbf{PMRC}} \\
\cline{2-9}
& \textbf{lr}
& \textbf{lr}
& \textbf{lr}
& \textbf{lr}
& \textbf{lr}
& \textbf{lr}
& \textbf{lr}
& $\gamma$ \\
\hline
LLaMA-$60$M
& $3\times 10^{-3}$
& $3\times 10^{-3}$
& $3\times 10^{-3}$
& $5\times 10^{-3}$
& $2\times 10^{-3}$
& $3\times 10^{-3}$
& $2\times 10^{-3}$
& $8192$ \\
LLaMA-$150$M
& $2\times 10^{-3}$
& $3\times 10^{-3}$
& $5\times 10^{-3}$
& $5\times 10^{-3}$
& $5\times 10^{-3}$
& $5\times 10^{-3}$
& $2\times 10^{-3}$
& $16384$ \\
LLaMA-$350$M
& $1\times 10^{-3}$
& $2\times 10^{-3}$
& $5\times 10^{-3}$
& $5\times 10^{-3}$
& $5\times 10^{-3}$
& $3\times 10^{-3}$
& $2\times 10^{-3}$
& $16384$ \\
LLaMA-$1.3$B
& $1\times 10^{-3}$
& $1\times 10^{-3}$
& $1\times 10^{-3}$
& $1\times 10^{-3}$
& $1\times 10^{-3}$
& $1\times 10^{-3}$
& $5\times 10^{-3}$
& $16384$ \\
\hline
\end{tabular}%
}
\end{table}

\begin{table}[h]
    \centering
\caption{Zero-shot accuracy (\%) on seven benchmarks. \textbf{Bold} and \underline{underlined} values indicate the best and second-best average accuracies among memory-efficient methods, respectively. Averages are computed from unrounded scores.}
    \label{tab:llm_pretraining_results}
     \setlength{\tabcolsep}{1.5pt}
    \renewcommand{\arraystretch}{1.12}
    \begin{tabular}{c|ccccccccc}
        \hline
        Model
        & Optimizer
        & \textbf{ARC-E}
        & \textbf{ARC-C}
        & \textbf{OBQA}
        & \textbf{HellaSwag}
        & \textbf{PIQA}
        & \textbf{SIQA}
        & \textbf{WinoGrande}
        & \textbf{Avg.} \\
        \hline
        & AdamW & $33.67$ & $17.92$ & $12.80$ & $27.15$ & $59.74$ & $35.98$ & $51.07$ & $34.05$ \\
        & Muon  & $35.77$ & $17.83$ & $13.20$ & $27.46$ & $60.66$ & $35.82$ & $49.80$ & $34.37$ \\
        \cline{2-10}
        & GaLore & $34.39$ & $18.60$ & $14.40$ & $26.94$ & $58.87$ & $34.75$ &$52.33$ & $\underline{34.32}$ \\
        \raisebox{-0.5\height}[0pt][0pt]{%
            \shortstack[c]{LLaMA\\-$60$M}}
        & GoLore & $32.79$ & $17.75$ & $12.20$ & $27.03$ & $57.83$ & $37.10$ & $52.01$ & $33.82$ \\
        & Fira   & $34.93$ & $18.86$ & $12.20$ & $26.98$ & $59.19$ & $35.47$ & $50.75$ & $34.05$ \\
        & GUM   & $35.02$ & $18.34$ & $13.40$ & $27.04$ & $58.27$ & $35.98$ & $51.62$ & $34.23$ \\
        & \textbf{PMRC} & $34.89$ & $18.43$ & $13.40$ & $27.32$ & $59.41$ & $36.69$ & $52.41$ & $\textbf{34.65}$ \\
        \hline
        & AdamW & $38.59$ & $17.49$ & $14.60$ & $28.39$ & $61.64$ & $35.67$ & $50.83$ & $35.32$ \\
        & Muon  & $40.03$ & $17.66$ & $16.40$ & $28.64$ & $62.73$ & $36.80$ & $50.43$ & $36.10$ \\
        \cline{2-10}
        & GaLore & $38.22$ & $17.58$ &  $13.60$ & $27.99$ & $60.61$ & $37.10$ & $50.51$ & $35.09$ \\
        \raisebox{-0.5\height}[0pt][0pt]{%
            \shortstack[c]{LLaMA\\-$150$M}}
        & GoLore & $38.13$ & $16.38$ & $15.00$ & $27.76$ &  $60.88$ & $36.34$ & $50.99$ & $35.07$ \\
        & Fira   & $38.59$ & $18.17$ & $13.40$ & $28.15$ & $61.64$ & $35.88$ & $50.83$ & $35.24$ \\
        & GUM & $38.09$ & $18.34$ & $14.60$ & $27.89$ & $60.77$ & $37.15$ & $50.75$ &  $\underline{35.37}$ \\
         & \textbf{PMRC} & $38.76$ & $18.77$ & $15.80$ & $28.81$ & $61.21$ & $36.80$ & $52.17$ & $\textbf{36.05}$ \\
        \hline
        & AdamW & $42.68$ & $18.00$ & $16.80$ & $30.14$ & $64.96$ & $37.15$ & $52.80$ & $37.51$ \\
        & Muon  & $44.19$ & $19.97$ & $18.80$ & $32.08$ & $66.21$ & $38.02$ & $54.70$ & $39.14$ \\\cline{2-10}
        & GaLore & $40.03$ & $19.11$ & $15.20$ & $29.13$ & $62.95$ & $36.85$ & $51.62$ & $36.41$ \\
        \raisebox{-0.5\height}[0pt][0pt]{%
            \shortstack[c]{LLaMA\\-$350$M}}
        & GoLore & $39.86$ & $16.55$ & $15.00$ & $29.20$ & $63.66$ & $37.05$ & $50.20$ & $35.93$ \\
        & Fira   & $41.50$ & $17.58$ & $16.20$ & $29.94$ & $64.04$ & $36.85$ & $51.54$ & $36.81$ \\
        & GUM & $42.13$ & $19.54$ & $15.60$ & $30.15$ & $65.78$ & $37.31$ & $50.83$ & $\underline{37.33}$ \\
         & \textbf{PMRC} & $44.11$ & $20.14$ & $18.40$ & $31.25$ & $65.67$ & $38.54$ & $50.99$ & $\textbf{38.44}$ \\
        \hline

        & AdamW & $46.68$ & $21.25$ & $18.40$ & $34.79$ & $68.28$ & $39.15$ & $51.62$ & $40.02$ \\
        & Muon  & $51.60$ & $21.76$ & $19.60$ & $39.14$ & $71.00$ & $40.53$ & $51.14$ & $42.11$ \\
        \cline{2-10}
        & GaLore & $42.00$ & $19.45$ & $15.60$ & $29.93$ & $64.64$ & $37.41$ & $50.04$ & $37.01$ \\
        \raisebox{-0.5\height}[0pt][0pt]{%
            \shortstack[c]{LLaMA\\-$1.3$B}}
        & GoLore & $38.80$ & $17.66$ & $13.40$ & $28.90$ & $62.84$ & $34.80$ & $51.30$ & $35.39$ \\
        & Fira   & $45.03$ & $18.77$ & $16.20$  & $32.10$ & $67.25$ & $37.56$ & $51.85$ & $38.40$ \\
        &  GUM & $46.97$ & $20.14$ & $19.00$ & $33.62$ & $67.25$ & $39.71$ & $52.25$ & $\underline{39.85}$ \\
         & \textbf{PMRC} & $48.86$ & $20.65$ & $19.60$ & $36.01$ & $69.48$ & $39.76$ & $51.07$ & $\textbf{40.78}$ \\
        \hline
    \end{tabular}
\end{table}

\subsubsection{LLM Pretraining Tasks}\label{exp-pt}

We adopt the LLaMA architecture  and study four model scales with $60$M, $150$M, $350$M and $1.3$B parameters, whose architectural configurations are summarized in Table~\ref{tab:model_architecture}. We pretrain all models on the C4 dataset \citep{C4}. We implement the experiments  based on the code from \citet{glentis2025memory} \footnote{\url{https://github.com/OptimAI-Lab/Minimalist_LLM_Pretraining}} and calculate $\operatorname{msign}(\cdot)$ using  Newton–Schulz iterations. Our implementations of GaLore\footnote{\url{https://github.com/jiaweizzhao/galore}}, GoLore\footnote{\url{https://github.com/pkumelon/golore}} and Fira\footnote{\url{https://github.com/xichen-fy/Fira}} are based on their publicly released code. For AdamW-based methods, we set $\beta_1=0.9$ and $\beta_2=0.999$. For Muon-based methods, we set the momentum coefficient to $1-\beta=0.95$ and enable Nesterov momentum. We compute the matrix sign using $5$ Newton--Schulz iterations and optimize the one-dimensional and embedding-layer parameters with AdamW.
 
 We then evaluate the models on the following benchmarks: ARC-Easy, ARC-Challenge  \citep{clark2018think}, OpenBookQA \citep{mihaylov2018can}, HellaSwag \citep{zellers2019hellaswag}, PIQA \citep{bisk2020piqa}, SIQA \citep{sap2019social} and WinoGrande \citep{sakaguchi2021winogrande}. 

Each hyperparameter either follows the recommendation of the original paper or is selected by grid search. For each model and optimizer, we select the hyperparameter configuration that achieves the highest average accuracy across the seven downstream benchmarks, with the chosen values summarized in Table~\ref{tab:optimizer-hparams}. We report the memory cost of all methods in Table~\ref{tab:peak_gpu_memorypre}, measured with \texttt{torch.compile} disabled. As shown in Table~\ref{tab:llm_pretraining_results}, PMRC achieves the highest average accuracy among all memory-efficient methods across all model scales.

\begin{table}[h]
\centering
\caption{Accuracy (\%) on the IFEval (instruction-following) benchmark. \textbf{Bold} and \underline{underlined} values indicate the best and second-best results among memory-efficient methods, respectively.}
\label{tab:if}
\small
\renewcommand{\arraystretch}{1.4}
\setlength{\tabcolsep}{5pt}
\setlength{\arrayrulewidth}{0.6pt}

\resizebox{\textwidth}{!}{%
\begin{tabular}{cccccccc}
\hline
& & \multicolumn{3}{c}{\textbf{Prompt-level Strict Accuracy}}
& \multicolumn{3}{c}{\textbf{Instruction-level Strict Accuracy}} \\
\cline{3-5}\cline{6-8}

\textbf{Memory-Efficient}
& \textbf{Optimizer}
& \textbf{Qwen3-8B}
& \textbf{LLaMA-3.1-8B}
& \textbf{OLMo-2-1124-7B}
& \textbf{Qwen3-8B}
& \textbf{LLaMA-3.1-8B}
& \textbf{OLMo-2-1124-7B} \\
\hline
$\textcolor{red}{\times}$
& AdamW
& $56.38$ & $39.19$ & $26.80$ & $65.83$ & $50.12$ & $38.97$ \\
$\textcolor{red}{\times}$
& Muon
& $52.68$ & $34.38$ & $24.03$ & $62.11$ & $46.16$ & $36.45$ \\
\hline
$\textcolor{blue}{\checkmark}$
& GaLore
& $57.49$ & $\underline{36.60}$ & $\underline{26.62}$ & $66.67$ & $47.24$ & $\underline{38.13}$ \\
$\textcolor{blue}{\checkmark}$
& GoLore
& $\underline{58.96}$ & $32.90$ & $24.03$ & $\underline{68.35}$ & $\underline{47.96}$ & $35.73$ \\
$\textcolor{blue}{\checkmark}$
& GUM ($r=128$)
&  $51.76$ & $33.46$ & $24.95$ & $62.47$ & $43.65$ & $36.21$ \\
$\textcolor{blue}{\checkmark}$
& GUM ($r=256$)
& $55.45$ & $35.30$ & $26.25$ & $65.35$ & $46.76$ & $36.69$ \\
$\textcolor{blue}{\checkmark}$
& \textbf{PMRC}
& $\mathbf{59.89}$ & $\mathbf{37.52}$ & $\mathbf{27.36}$ & $\mathbf{69.06}$ & $\mathbf{50.24}$ & $\mathbf{38.97}$ \\
\hline
\end{tabular}%
}
\end{table}

\subsubsection{LLM Finetuning Tasks}\label{app:sft}

In this section, we provide the omitted details of the LLM supervised finetuning (SFT) experiments. For the instruction-following task, we train on a single NVIDIA RTX PRO 6000 GPU and adopt a device micro-batch size of $16$ with $8$ gradient accumulation steps for Muon, GaLore, GoLore, GUM, and PMRC, and a device micro-batch size of $8$ with $16$ gradient accumulation steps for AdamW, so that the global batch size is $128$ in both cases. For the mathematical reasoning task, we train on $4$ NVIDIA H20 GPUs and adopt a device micro-batch size of $4$ with $8$ gradient accumulation steps for the same methods and a device micro-batch size of $2$ with $16$ gradient accumulation steps for AdamW, so that the global batch size is again $128$.  For AdamW-based methods, we use the standard coefficients $\beta_1=0.9$ and $\beta_2=0.999$. For Muon-based methods, the momentum coefficient is set to $1-\beta=0.95$ with Nesterov momentum enabled. The matrix sign is approximated using $5$ Newton--Schulz iterations, while one-dimensional parameters and embedding layers are optimized with AdamW.

\textbf{Hyperparameter Tuning.}
 For each model and optimization method we select the learning rate by a grid search over $\{1\times10^{-5},2\times10^{-5}, 4\times10^{-5}, 8\times 10^{-5}\}$ for the instruction-following task  and $\{8\times 10^{-6}, 1\times10^{-5},3\times10^{-5},5\times10^{-5},7\times10^{-5}\}$ for the mathematical-reasoning task. PMRC uses a residual cap at a ratio of $4$ and we grid search $\gamma \in \{2000,4000,8000,16000\}$. We select the parameters that achieve the highest accuracy on benchmarks.

\textbf{Additional results.} We present the prompt-level and instruction-level strict accuracy on the IFEval benchmark in Table~\ref{tab:if}. For completeness, we include the results of GUM with rank $r=256$, in addition to the memory-matched setting with $r=128$ used in the main experiments. PMRC outperforms GUM at both ranks on all three models in both metrics. We also report the peak allocated GPU memory of all methods in Table~\ref{tab:peak_gpu_memory}, measured on a single NVIDIA H20 GPU with \texttt{torch.compile} disabled. PMRC requires the lowest peak memory across all models.


\begin{table}[t]
    \centering
    \caption{Peak allocated GPU memory (GiB) during full-parameter SFT. We use a micro-batch size of $16$ with $8$ gradient accumulation steps for IFEval and a micro-batch size of $4$ with $32$ gradient accumulation steps for GSM8K. All memory measurements are performed on a single H20 GPU.}
    \label{tab:peak_gpu_memory}
    \renewcommand{\arraystretch}{1.4}
    \resizebox{0.9\linewidth}{!}{%
        \begin{tabular}{cccccccc}
            \hline
            \textbf{Task} & \textbf{Model} & \textbf{AdamW} & \textbf{Muon} & \textbf{GaLore} & \textbf{GoLore} & \textbf{GUM} & \textbf{PMRC} \\
            \hline
            \multirow{3}{*}{\textbf{IFEval}}
                & Qwen3-8B     & $89.16$  & $76.23$ & $67.71$  & $67.71$ & $66.01$ & $\textbf{65.88}$ \\
                & LLaMA-3.1-8B & $78.32$ & $65.32$ & $56.63$ & $56.63$ & $54.90$  & $\textbf{54.81}$ \\
                & OLMo-2-1124-7B   & $69.74$ & $57.43$ & $49.23$ & $49.23$ & $47.79$ & $\textbf{47.75}$ \\
            \hline
            \multirow{3}{*}{\textbf{GSM8K}}
                & Qwen3-8B     & $79.46$ & $66.52$ &  $58.02$ & $58.02$ & $57.00$ & $\textbf{56.18}$ \\
                & LLaMA-3.1-8B & $74.98$ & $61.98$ & $53.30$ & $53.30$ & $52.35$ & $\textbf{51.48}$ \\
                & OLMo-2-1124-7B   & $66.77$ & $54.71$ & $46.54$ & $46.54$ & $45.80$ & $\textbf{45.03}$ \\
            \hline
        \end{tabular}%
    }
\end{table}


\begin{figure}[t!]
\centering
\includegraphics[draft=false,width=0.3\textwidth]{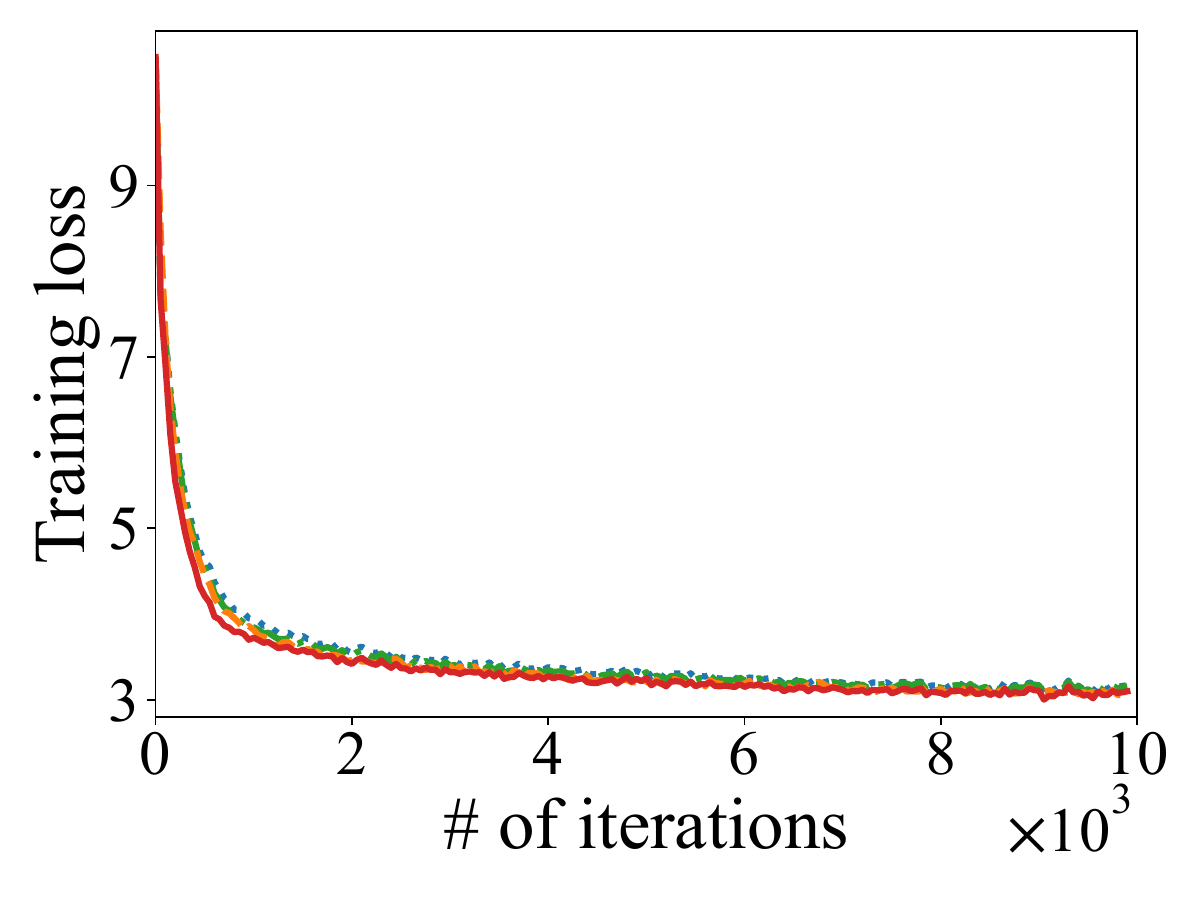}
\includegraphics[draft=false,width=0.3\textwidth]{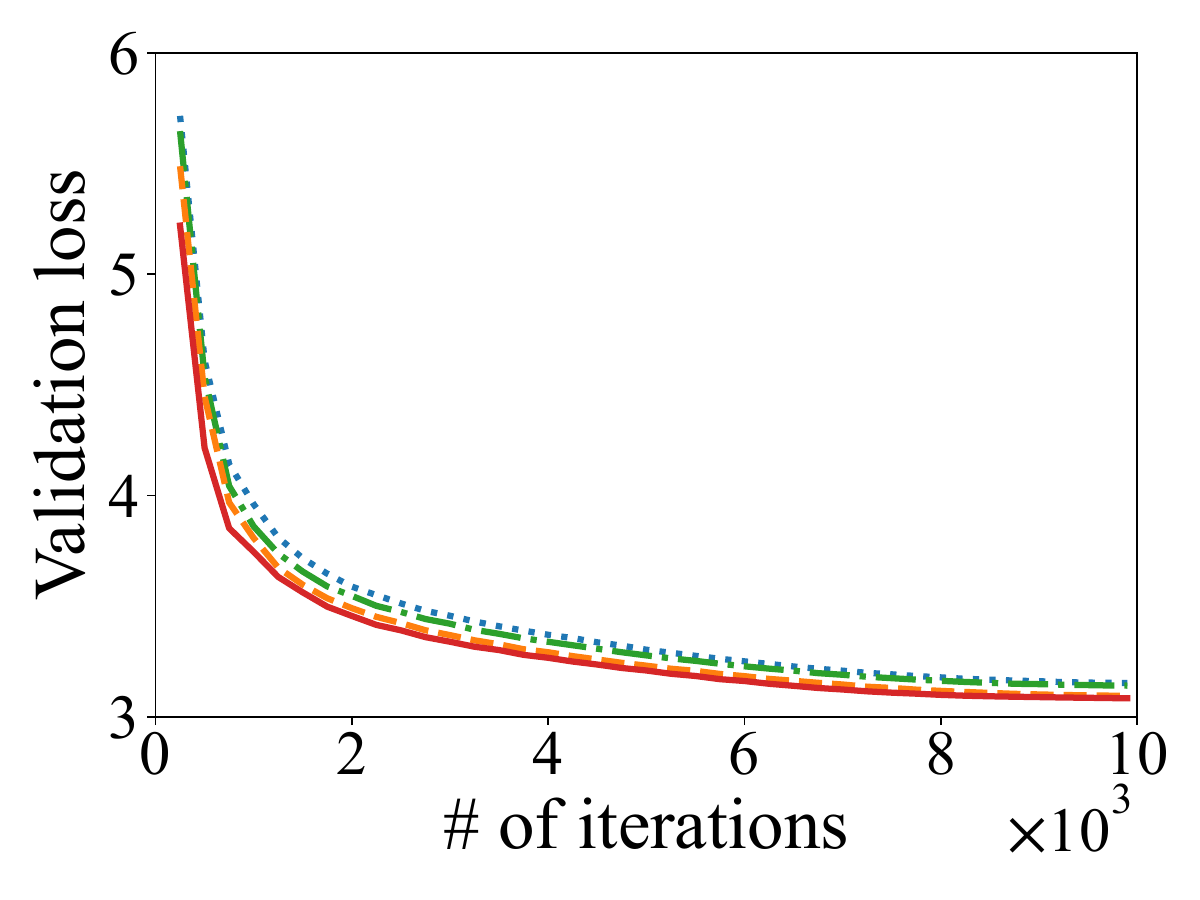}\\
\includegraphics[draft=false,width=0.6\textwidth]{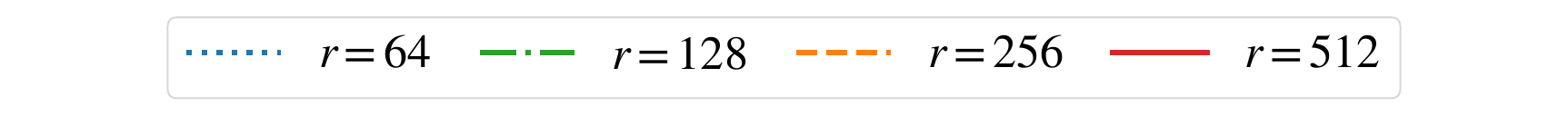}
\caption{Effect of the subspace rank $r$ on PMRC.}
\label{fig:rank-ablation}
\end{figure}

\begin{figure}[t!]
\centering
\includegraphics[draft=false,width=0.3\textwidth]{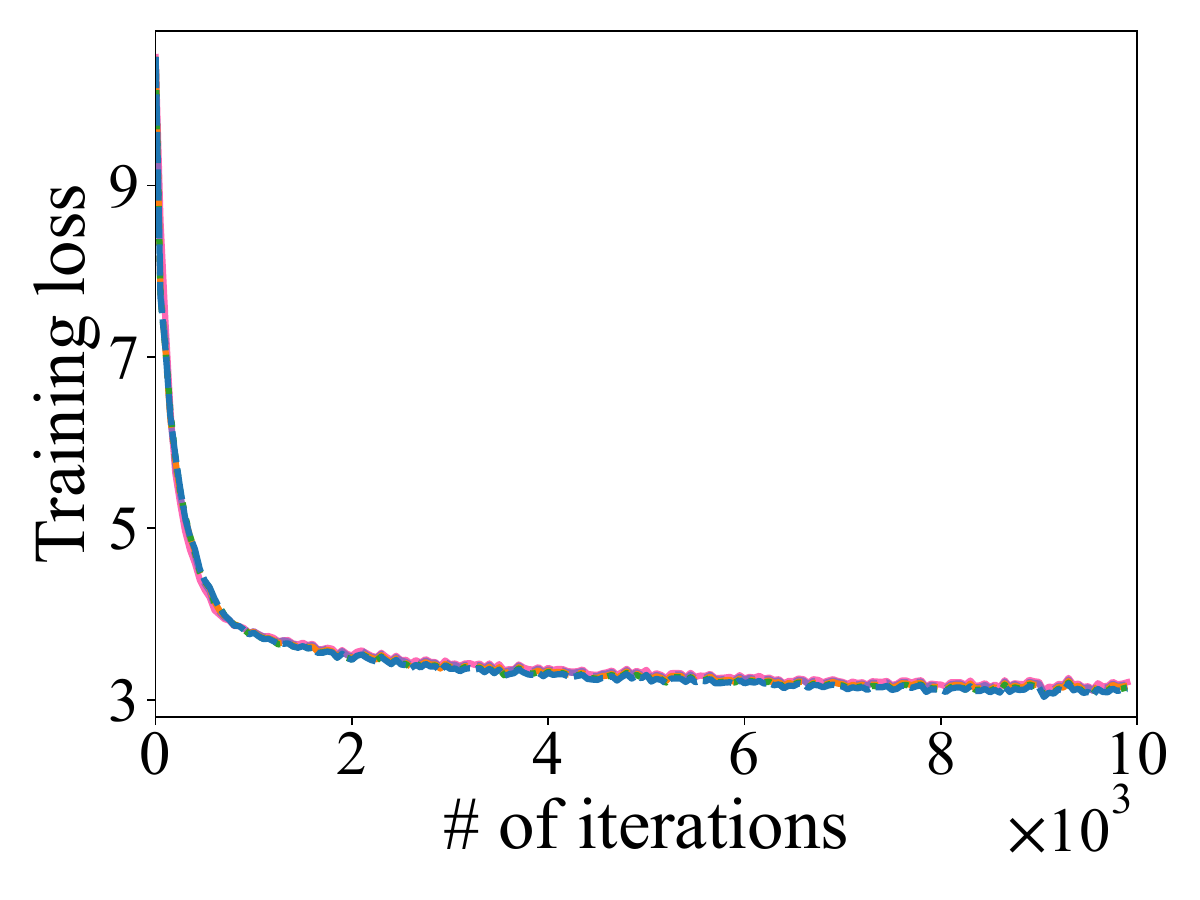}
\includegraphics[draft=false,width=0.3\textwidth]{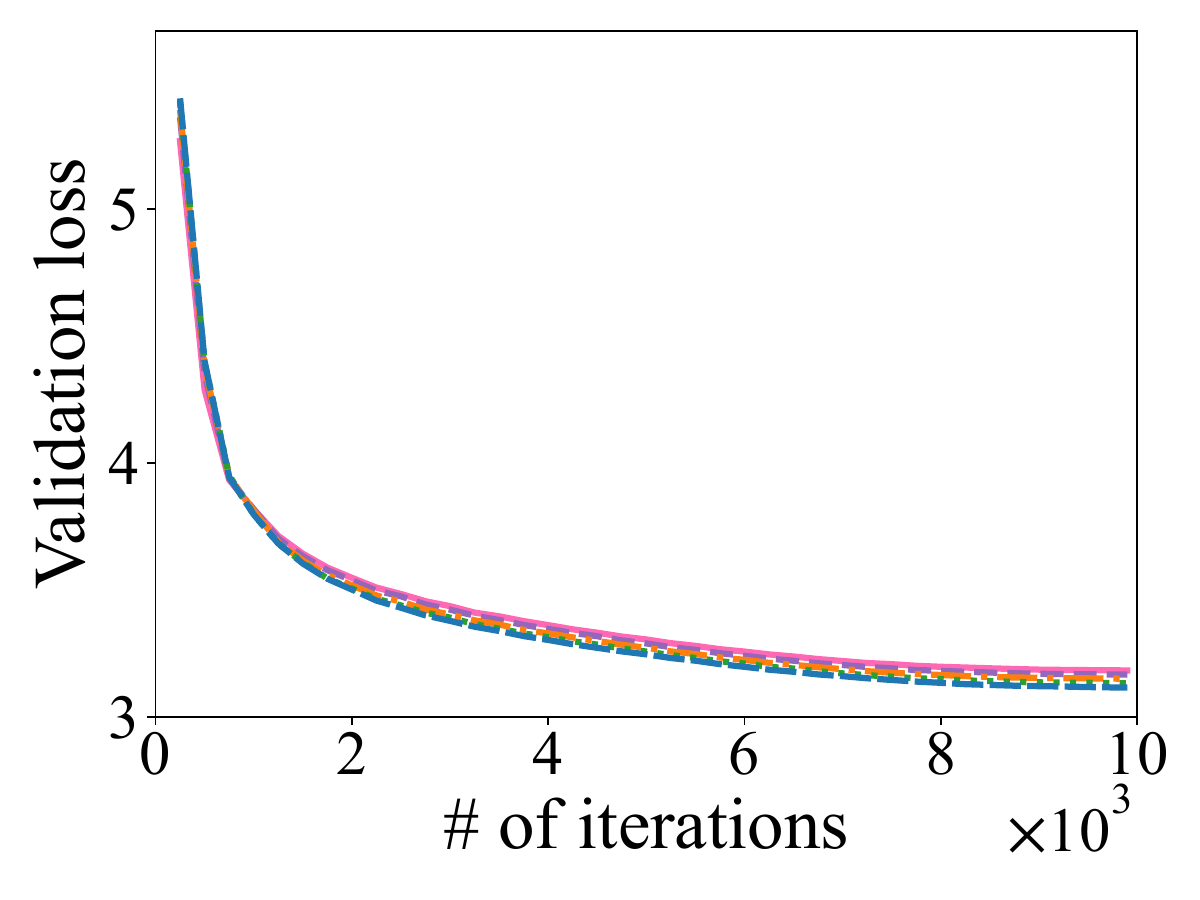}\\
\includegraphics[draft=false,width=0.6\textwidth]{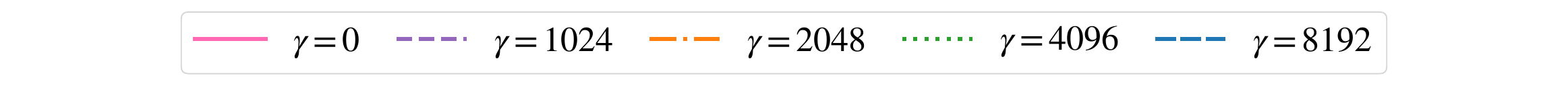}
\caption{Effect of the residual weight $\gamma$ on PMRC.}
\label{fig:gamma-ablation}
\end{figure}

\subsubsection{Ablation Studies}\label{abla}

For all ablations, we pretrain LLaMA-150M on the C4 dataset for $9900$ steps with a sequence length of $512$ and a global batch size of $512$. We refresh the projection basis every $100$ steps and, unless otherwise stated, set the residual cap to $\rho=4$.

\textbf{Effect of the subspace rank.}
We vary the projection rank over $r\in\{64,128,256,512\}$ with the residual weight fixed at $\gamma=8192$ and all other settings unchanged. Increasing $r$ expands the subspace in which momentum is maintained and the matrix-sign transformation is applied. As shown in Figure~\ref{fig:rank-ablation}, a larger rank reduces the validation loss, but also increases the memory required to store the projection basis and the projected momentum. For a parameter matrix of size $m\times n$, this memory cost scales as $O((m+n)r)$. These results illustrate the trade-off between validation performance and optimizer-state memory when choosing the subspace rank.

\textbf{Effect of the residual weight.}
We fix the rank at $r=256$ and vary the residual weight over $\gamma\in\{0,1024,2048,4096,8192\}$ with all other settings unchanged. Since the residual cap remains active, $\gamma$ serves as an upper bound on the effective weight applied to each parameter matrix rather than its exact value. As shown in Figure~\ref{fig:gamma-ablation}, the validation loss decreases monotonically as $\gamma$ increases over the tested range, with diminishing improvements at larger values. In particular, $\gamma=0$ removes the residual correction and yields the highest validation loss. This comparison supports the benefit of incorporating gradient information from the orthogonal complement in LLM pretraining.

\begin{figure}[t!]
\centering
\includegraphics[draft=false,width=0.3\textwidth]{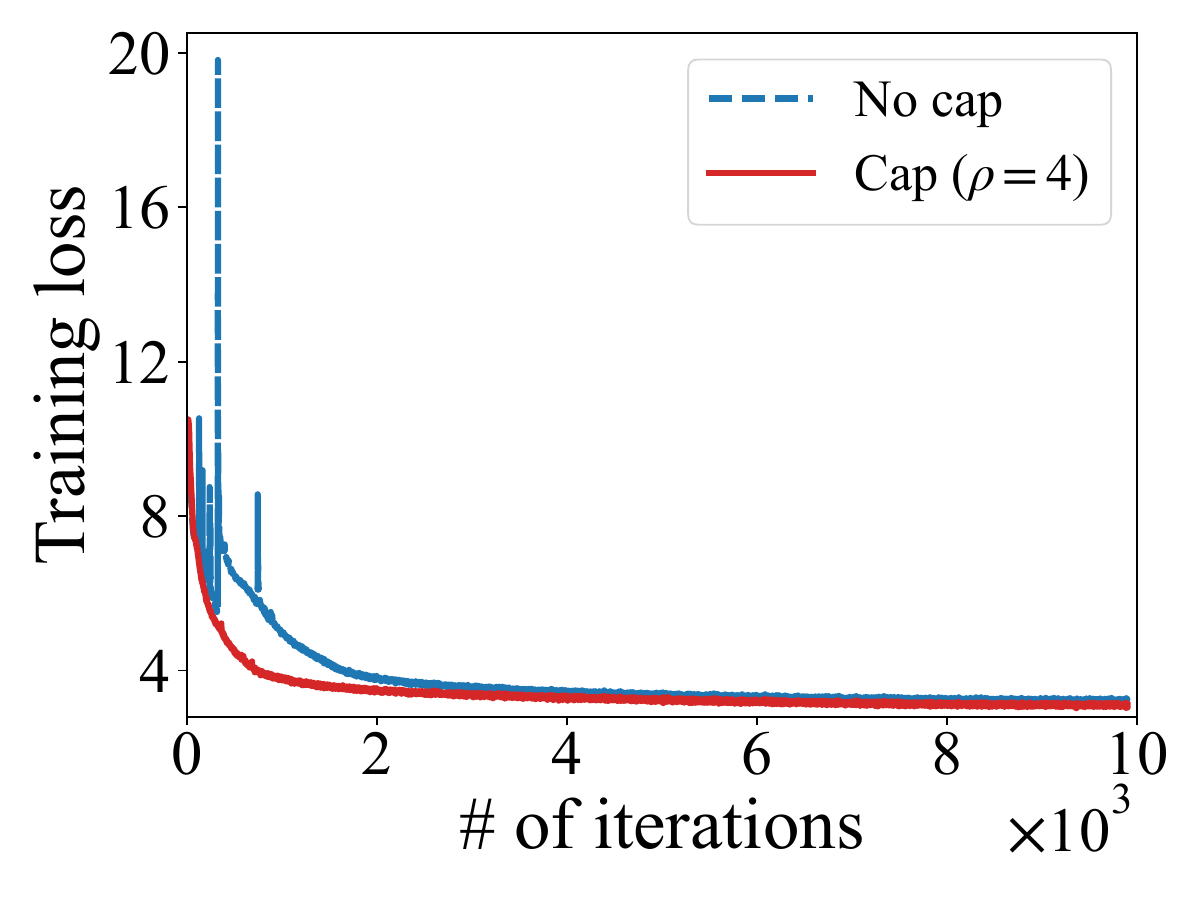}
\caption{Effect of the residual cap on PMRC.}
\label{fig:cap-ablation}
\end{figure}

\textbf{Effect of the residual cap.}
We finally examine the role of the residual cap in stabilizing training. Under the practical update in \eqref{finalupdate}, the effective residual weight is
\begin{equation*}
\widehat{\gamma}_{k,t}
=
\min\left\{
\gamma,
\rho\sqrt{
\frac{\lVert P_kZ_{k,t}\rVert_F^2}
{\max\{\lVert R_{k,t}\rVert_F^2,\epsilon\}}
}
\right\}.
\end{equation*}
By construction, this rule ensures
$\lVert\widehat{\gamma}_{k,t}R_{k,t}\rVert_F
\leq \rho\lVert P_kZ_{k,t}\rVert_F$. We compare the default choice $\rho=4$ with an uncapped run, fixing $r=256$ and $\gamma=8192$ in both runs. Without the cap, the full residual gradient is multiplied by $\gamma$ at every step. As shown in Figure~\ref{fig:cap-ablation}, the uncapped run becomes unstable during early training. This observation supports the use of the residual cap to control excessively large residual updates, while retaining the original coefficient $\gamma$ whenever the correction is below the threshold.

\end{document}